\documentclass[10pt,twocolumn,letterpaper]{article}

\usepackage{cvpr}
\usepackage{times}
\usepackage{epsfig}
\usepackage{graphicx}
\usepackage{amsmath}
\usepackage{amssymb}
\usepackage{enumitem}
\usepackage{float}
\usepackage{lipsum}
\usepackage{capt-of,etoolbox}
\usepackage{mwe}
\usepackage{xcolor}

\def \saliency {\textup{\saliency}}

\def \path {\mathit{path}}

\usepackage{algorithm}
\usepackage{algorithmic}

\usepackage[pagebackref=true,breaklinks=true,letterpaper=true,colorlinks,bookmarks=false]{hyperref}

\cvprfinalcopy 

\def\cvprPaperID{6560} 

\ifcvprfinal\fi

\begin{document}

\title{Panoptic-based Image Synthesis}

\author{Aysegul Dundar \qquad Karan Sapra \qquad Guilin Liu \qquad Andrew Tao \qquad Bryan Catanzaro \\
 NVIDIA 
}

\makeatletter
\let\@oldmaketitle\@maketitle
\renewcommand{\@maketitle}{\@oldmaketitle
\centering
\includegraphics[width=0.9\textwidth]{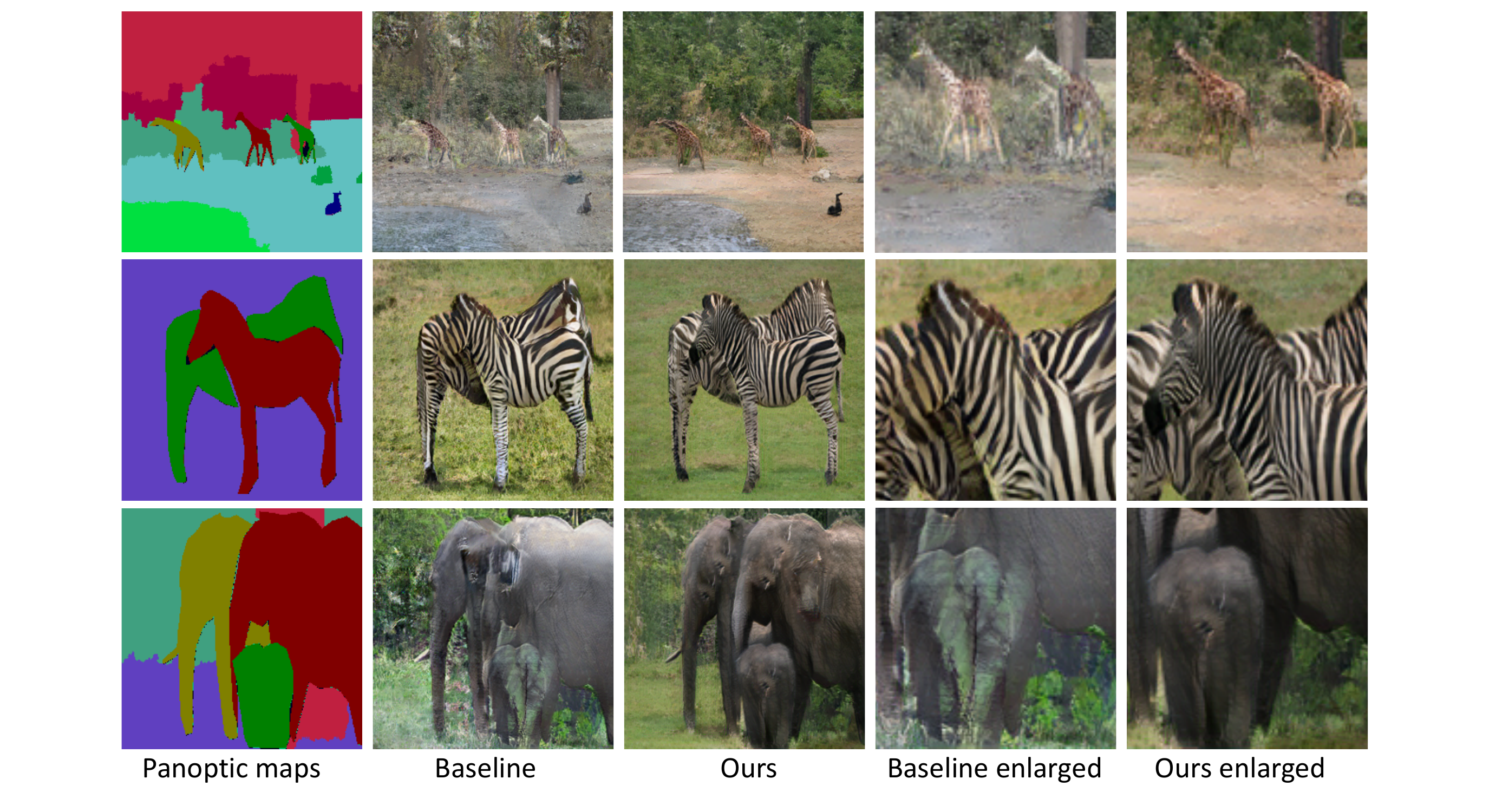}
\captionof{figure}{Unlike previous methods that rely on semantic and boundary label maps to synthesize images, our model uses panoptic maps. It generates instances with clear separation  even in cluttered scenes where multiple instances occlude each other. The forth and fifth column show zoomed in patches from the second and third column images to highlight the boundaries between instances, where previous methods tend to blend instances together.
}
\label{fig:teaser}
  \bigskip
  \medskip
  }
\makeatother

\maketitle


\begin{abstract}
Conditional image synthesis for generating photorealistic images serves various applications for content editing to content generation. Previous conditional image synthesis algorithms mostly rely on semantic maps, and often fail in complex environments where multiple instances occlude each other. We propose a panoptic aware image synthesis network to generate high fidelity and photorealistic images conditioned on  panoptic maps which unify semantic and instance information. To achieve this, we efficiently use panoptic maps in convolution and upsampling layers. We show that with the proposed changes to the generator, we can improve on the previous state-of-the-art methods by generating images in complex instance interaction environments in higher fidelity and tiny objects in more details. Furthermore, our proposed method also outperforms the previous state-of-the-art methods in metrics of mean IoU (Intersection over Union), and detAP (Detection Average Precision). 
\end{abstract}

\section{Introduction}

Image synthesis refers to the task of generating diverse and photo-realistic images, where a prevalent sub-category known as conditional image synthesis outputs images that are conditioned on some input data. Recently, deep neural networks have been  successful at conditional image synthesis ~\cite{isola2017image,chen2017photographic,wang2018high, zhang2017stackgan,zhang2017stackgan++,zhang2018self,brock2018large} where one of the  conditional inputs is a semantic segmentation map.
Extending this concept, in this paper, we are interested in the generation of photo-realistic images guided by panoptic maps.
Panoptic maps unify semantic and instance maps.
Specifically, they provide information about object instances for countable classes which are called ``things"  such as people, animals, and cars.
Additionally, they contain the semantic information about classes that are amorphous regions and repeat patterns or textures such as grass, sky, and wall. These classes are referred to as ``stuff".

We are interested in panoptic maps because semantic maps do not provide sufficient information to synthesize ``things" (instances)  especially in complex environments with multiple of them interacting with each other.
Even the state-of-the-art baseline (SPADE ~\cite{park2019semantic}) which inputs boundary maps to the network fails to generate high fidelity images when objects are small and instances are partially occluding.
This issue can be observed in Figure ~\ref{fig:teaser}, in the continuous pattern extending from one zebra to the other.
This is the result of conventional convolution and upsampling algorithms being independent of class and instance boundaries. To address this problem, we replace the convolution and upsampling layers in the generator with \textit{Panoptic aware convolution} and \textit{Panoptic aware upsampling} layers.

We refer to this form of image synthesis as panoptic-based image synthesis.
We evaluate our proposed image generator on two diverse and challenging datasets: Cityscapes \cite{cordts2016cityscapes}, and COCO-Stuff \cite{caesar2018coco}.
We demonstrate that we are able to efficiently and accurately use panoptic maps to generate higher fidelity images and improve on evaluation metrics used by previous methods ~\cite{chen2017photographic,wang2018high,park2019semantic}. 

Our main contributions can be summarized as follows:
\begin{enumerate}
    \item We propose to use \textit{Panoptic aware convolution} that re-weights convolution based on the panoptic maps in conditional image generation setting. Similar mechanisms have been previously used for other tasks \cite{liu2018image, harley2017segmentation} with binary masks and learned soft masks but not for image synthesis with multi-class panoptic masks. 
    \item We propose \textit{Panoptic aware upsampling} that addresses the misalignment between the upsampled low resolution features and high resolution panoptic maps. This ensures that the semantic and instance details are not lost, and that we also maintain higher accuracy alignment between the generated images and the panoptic maps.
    \item We demonstrate that using our proposed network architecture, not only do we see more photorealistic images, but we also observe significant improvements in object detection scores on both Cityscapes and COCO-Stuff datasets when evaluated with object detection model. 

\end{enumerate}


\section{Related Work}

\textbf{Generative Adversarial Networks (GANs)}
 ~\cite{goodfellow2014generative} perform image synthesis by modelling the natural image distribution and synthesizing new samples that are indistinguishable from natural images. This is achieved by using a generator and a discriminator network that are both trying to optimize an opposing objective function, in a zero-sum game.
 Many conditional image synthesis works use GANs to generate realistic images, and so does ours.

\textbf{Conditional Image Synthesis} can vary based on different type of inputs to be conditioned upon.
For example, inputs can be text ~\cite{reed2016generative,zhang2017stackgan,xu2017attngan,hong2018inferring}, natural and synthetic images ~\cite{karacan2016learning,zhu2017unpaired,liu2017unsupervised,zhu2017toward,huang2018multimodal,Zhao2018layout,karacan2018manipulating}, or unsupervised landmarks \cite{lorenz2019unsupervised,jakab2018unsupervised, shih2019video, dundar2020unsupervised} to name a few.
Recently, \cite{qi2018semi, chen2017photographic, isola2017image} use semantic maps and ~\cite{wang2018high, park2019semantic} use both semantic maps and boundary maps as inputs to the generator, where the boundary maps are obtained from the instance maps.
A pixel in the boundary map is set to 1 if its object identity is different from any of its 4 neighbors, and set to 0 otherwise.
This approach does not preserve the whole information contained in an instance map especially when instances are occluding each other.
The pixels that belong to the same instance may be separated by multiple boundaries.

\textbf{Content Aware Convolution.} There have been many works that learn to weight the convolution activations based on attention mechanisms \cite{zhang2018self, wang2018non, yang2016stacked, harley2017segmentation}. These mechanisms operate on feature maps to capture the spatial locations that are related to each other while making a decision.
In another line of research, the spatial locations that should not contribute to an output may be given to us by binary masks such as in the case of image inpainting, the task of filling in holes in an image.
In this task, ~\cite{liu2018partial, uhrig2017sparsity} use partial convolutions so that given a binary mask with holes and valid pixels, the convolutional results depend only on the valid pixels. 
Our convolution layer is similar to the one used in image inpainting, instead of masks with holes, we have panoptic maps given to us, and therefore we know that convolutional results of an instance should not depend on an another instance or on pixels that belong to a different semantic class.
We are not given binary masks, but we generate them efficiently on-the-fly based on panoptic maps.

\textbf{Content Aware Upsampling.}
Nearest neighbor and bilinear interpolations are the most commonly used upsampling methods in deep learning applications. These methods use hand-crafted algorithms based on the relative positions of the pixel coordinates. There has been also great interest in learning the upsampling weights for the tasks of semantic segmentation  ~\cite{Noh_2015, su2019pixel} and image and video super-resolution \cite{shi_2016}.
Recently, \cite{mazzini2018guided, wang2019carafe} proposed feature guided upsampling algorithms.
These methods operate on the feature maps to encode contents, and based on the contents they upsample the features.
In our method, similar to the idea in the panoptic aware convolution layer, we take advantage of the high resolution panoptic maps to resolve the misalignments in upsampled feature maps and panoptic maps.

\section{Method}

In this section, we first detail the \textit{Panoptic aware convolution} and \textit{Panoptic aware upsampling} layers. We then describe the overall network architecture.

\subsection{Panoptic Aware Convolution Layer}

\begin{figure}
    \centering
    \includegraphics[width=0.45\textwidth]{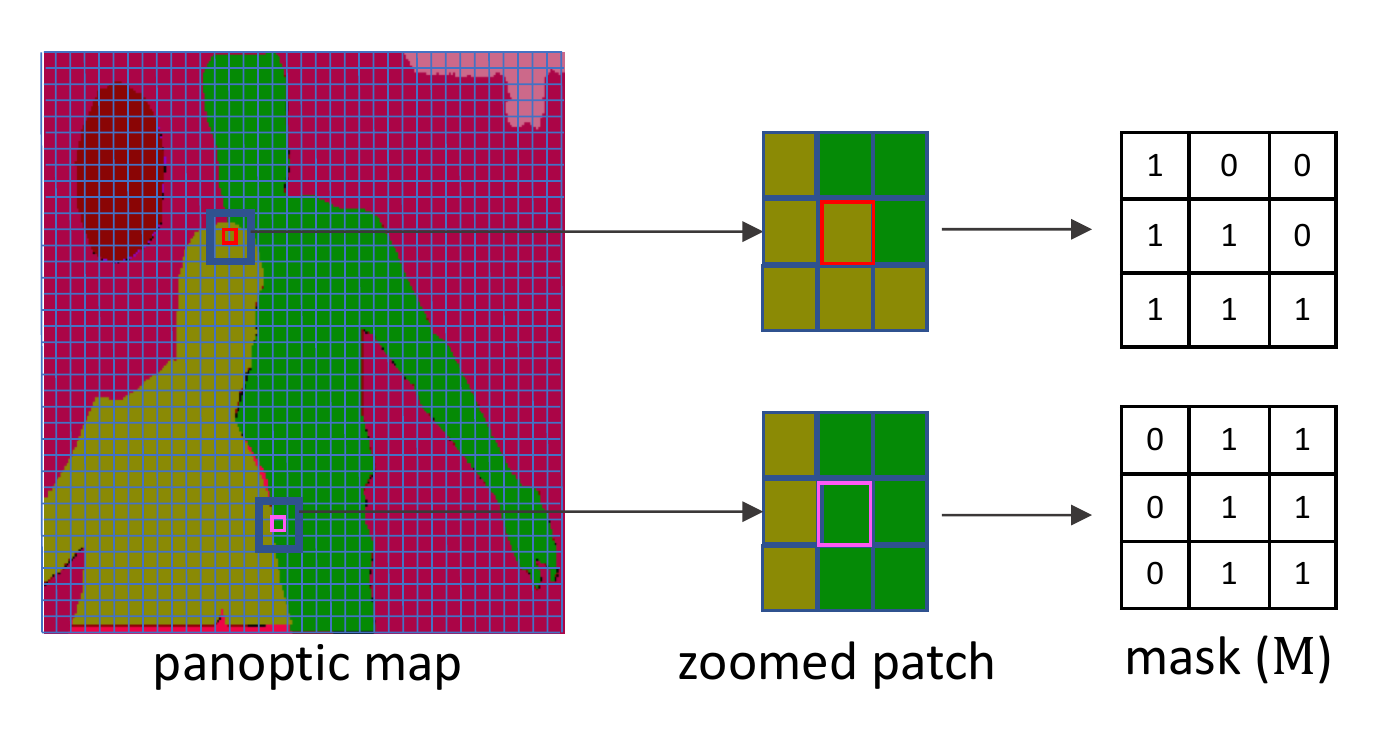}
    \caption{Panoptic aware partial convolution layer takes a panoptic map (colorized for visualization) and based on the center of each sliding window it generates a binary mask,  $\mathbf{M}$.
    The pixels that share the same identity with the center of the window are assigned 1 and the others 0.}
    \label{fig:partial_conv}
\end{figure}

We refer to the partial convolution operation using panoptic maps as a \emph{Panoptic aware partial convolution} layer which shares the fundamentals with other works that use partial convolution for different tasks \cite{harley2017segmentation, liu2018image}. Let $\mathbf{W}$ be the convolution filter weights and $b$ the corresponding bias. $\mathbf{X}$ is the feature values, $\mathbf{P}$ is the panoptic map values for the current convolution (sliding) window, and $\mathbf{M}$ is the corresponding binary mask. 

$\mathbf{M}$ defines which pixels will contribute to the output of the convolution operation based on  the panoptic maps. 
The pixel coordinates which share the same identity with the center pixel in the panoptic map are assigned 1 in the mask, while the others are assigned 0. This is expressed as:
\begin{equation}
m_{(i,j)}=\begin{cases}
    1, & \text{if } \mathbf{P_{(i, j)}==P_{(center, center)}}  \\
    0, & \text{otherwise}
\end{cases}
\end{equation}

This can be implemented by first subtracting the center pixel from the patch and clipping the absolute value to (0, 1), then subtracting the clipped output from 1 to inverse the zeros and ones.
Figure \ref{fig:partial_conv} depicts the construction of the mask  $\mathbf{M}$.

The partial convolution at every location is expressed as:

\begin{equation}
\label{eq:partconv}
    x' = \begin{cases}
     \mathbf{W}^T(\mathbf{X}\odot\mathbf{M})\frac{\text{sum}(\mathbf{1})}{\text{sum}(\mathbf{M})} + b,  & \text{if }  \text{sum}(\mathbf{M})>0  \\
        0,  & \text{otherwise}
    \end{cases}
\end{equation}

where $\odot$ denotes element-wise multiplication and $\mathbf{1}$ has same shape as $\mathbf{M}$ but with all elements being 1.  The scaling factor, $\text{sum}(\mathbf{1})/\text{sum}(\mathbf{M})$, applies normalization to account for varying amount of valid inputs as in \cite{liu2018image}.
With Equation \ref{eq:partconv}, convolution results of an instance or stuff depend only on the feature values that belong to the same instance or stuff.

\subsection{Panoptic Aware Upsampling Layer}
\begin{figure}
    \centering
    \includegraphics[width=0.45\textwidth, trim={0 1cm 0 0},clip]{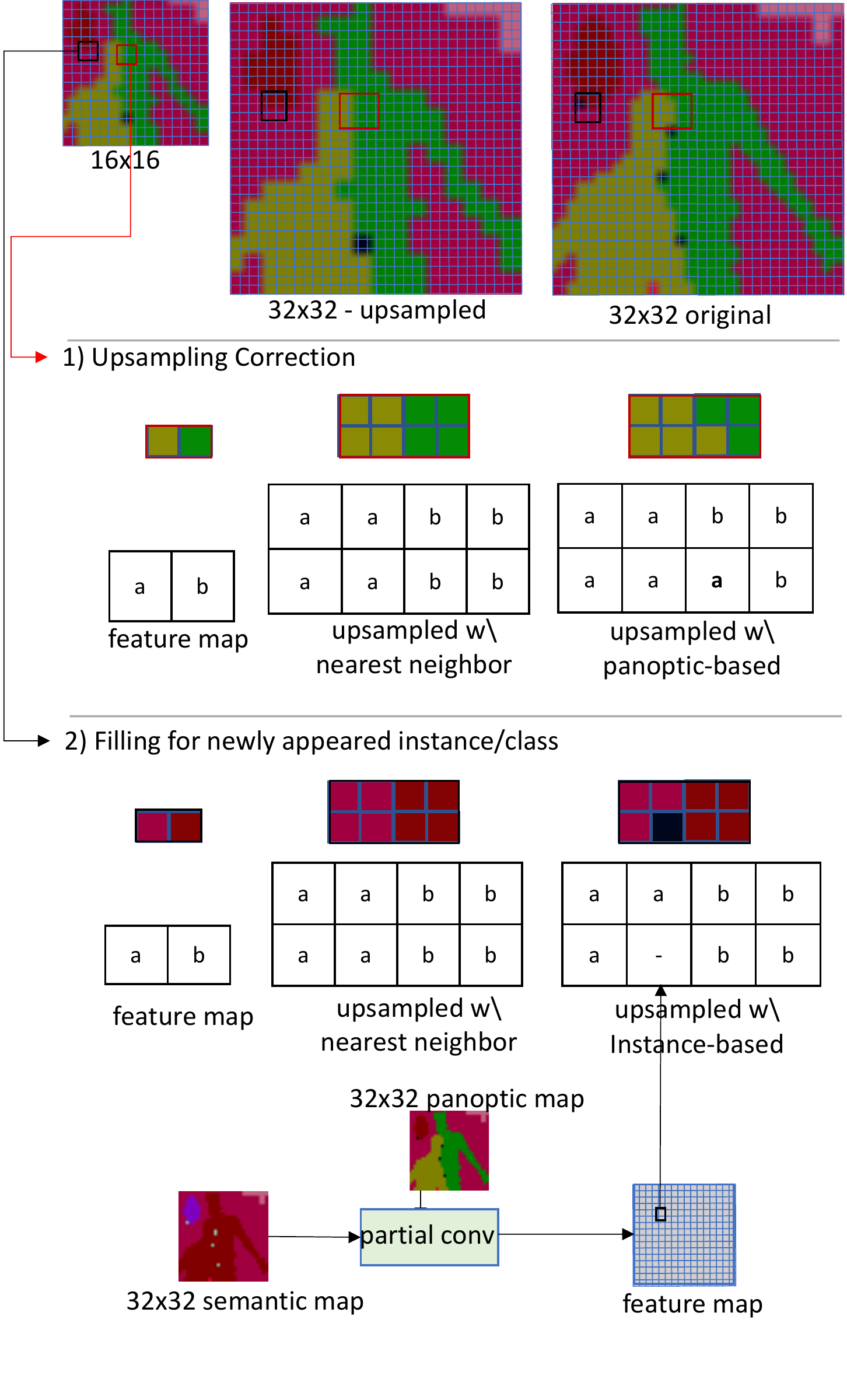}
    \caption{Overview of Panoptic aware upsampling module. $16\times16$ and $32\times32$ panoptic maps are nearest neighbor downsampled from original $256\times256$ panoptic map.  $32\times32$ upsampled map is upsampled from $16\times16$ panoptic map using nearest neighbor upsampling algorithm.
    Comparing the $32\times32$ upsampled and $32\times32$ original map, we can observe two issue: 1) Spatial misalignments and 2) Appearance of new classes or instances.
    As shown in Figure (top), first we correct misalignment by replicating a feature vector from a neighboring pixel that belongs to the same panoptic instance. This operation is different from nearest neighbor upsampling which would always replicate the top-left feature.
    Second, as shown in Figure (bottom), we resolve pixels where new semantic or instance classes have just appeared by encoding new features from semantic maps with \textit{Panoptic aware convolution} layer.}
    \label{fig:instance_upsample}
\end{figure}

We propose a \textit{Panoptic aware upsampling} layer as an alternative to traditional upsampling layers when the higher resolution panoptic maps are available as in the case of image synthesis for content generation task. Nearest neighbor upsampling is a popular conventional upsampling choice in conditional image synthesis tasks as used by \cite{brock2018large, park2019semantic, wang2018high, park2019semantic}.
However, nearest neighbor upsampling algorithm is hand-crafted to do replication. For example, in a $2\times2$ upsampling scenario, nearest neighbor algorithm will replicate the top-left corner to the neighboring pixels in a $2\times2$ window. This creates two issues as shown in Figure~\ref{fig:instance_upsample}. 

First, it can create a spatial misalignment between the high resolution panoptic map and upsampled features. Figure~\ref{fig:instance_upsample}-top illustrates this issue, where the features of instance $id_B$ are replicated and incorrectly used for instance $id_A$ following traditional upsampling approach.
In Figure~\ref{fig:instance_upsample}, we demonstrate the misalignments in the upsampled panoptic maps for clarity, but we are only interested in the alignment in feature maps.
We refer to the operation for fixing this misalignment as ``Upsampling alignment" correction.
Secondly, as shown in Figure~\ref{fig:instance_upsample}-bottom, the high resolution panoptic map may contain new classes and instances that might not exist in the lower resolution panoptic map. This implies that new features need to be generated and replaced in upsampled feature map. We refer to this  operation as ``Hole filling".

\begin{figure}[]
    \centering
    \includegraphics[width=0.5\textwidth]{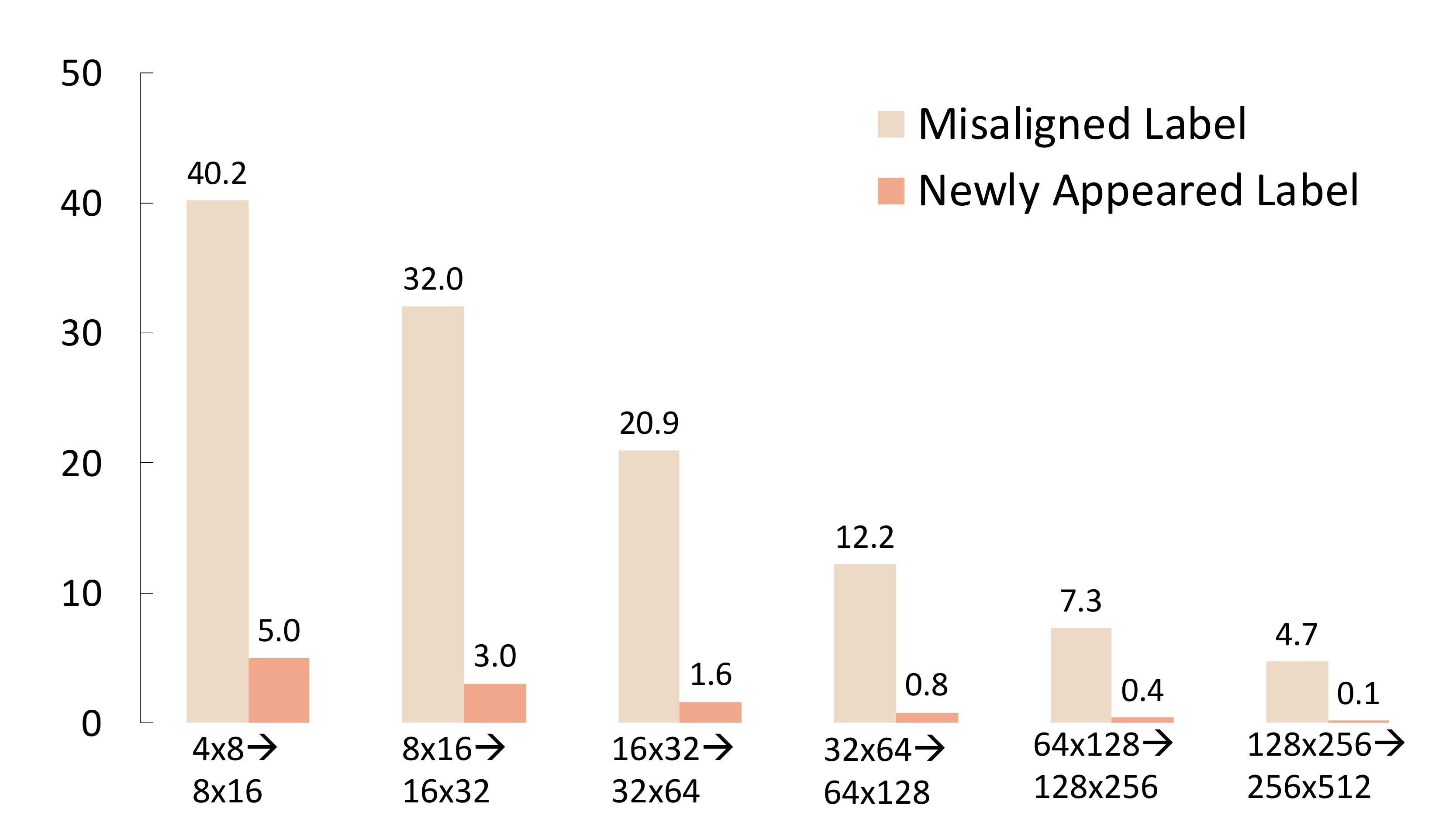}
    \caption{The percent of incorrectly mapped features using upsampling through different layers of network.}
    \label{fig:stats_upsample}
    
\end{figure}

Figure \ref{fig:stats_upsample} depicts how often the two issues mentioned above occur at different layers in the network for the Cityscapes dataset. As seen in the figure, especially in the early layers, over  $30\%$ of pixel features among the newly generated ones do not align with the panoptic maps, and many pixels that belong to a new instance or semantic map appear for the first time at the new scale.

To resolve these two issues, ~\textit{Panoptic aware upsampling} layer performs a two-step process: Upsample alignment correction and Hole filling as shown in Figure~\ref{fig:instance_upsample}.
Let $\mathbf{S}$ be the semantic map, and $\mathbf{F}$ the feature to be upsampled.
We are interested in $2\times2$ upsampling as it is the most common upsampling scale used by image synthesis methods.
Let $\mathbf{P^d}$ be the downsampled panoptic mask. We are interested in upsampling $\mathbf{F^d}$, to generate the  upsampled feature map, $\mathbf{F'^u}$ with the guidance from a higher scale panoptic and semantic maps, $\mathbf{P^u}$, and $\mathbf{S^u}$, and a  mask, $\mathbf{M^{correction}}$,  that we will generate.

\begin{algorithm}[tb!]
  \caption{Upsampling Alignment Correction.}
  \label{alg:misalign}
\begin{algorithmic}
\begin{small}
  \STATE {\bfseries Initialize:}
  $M^{correction} = 0$,
  $F'^u = 0$,
 \FOR{$i \in [0, 2W)$; $j \in [0,2H)$}
  \IF{ $P_{i, j}^u  == P_{i//2, j//2}^d$ }
    \STATE {$ {F'}_{i, j}^u = F^{d}_{i//2, j//2} $}
    \STATE $M^{correction}_{i, j} = 1$
  \ENDIF
 \ENDFOR
 \FOR{$i \in [0, 2W)$; $j \in [0,2H)$}
  \IF{ $P_{i, j}^u  == P_{i//2+1, j//2}^d$ \AND $M^{correction}_{i, j} != 1$}
    \STATE {$ {F'}_{i, j}^u = F^{d}_{i//2+1, j//2} $}
    \STATE $M^{correction}_{i, j} = 1$
  \ENDIF
   \ENDFOR
 \FOR{$i \in [0, 2W)$; $j \in [0,2H)$}
  \IF{ $P_{i, j}^u  == P_{i//2, j//2+1}^d$ \AND $M^{correction}_{i, j} != 1$}
    \STATE {$ {F'}_{i, j}^u = F^{d}_{i//2, j//2+1} $}
    \STATE $M^{correction}_{i, j} = 1$
  \ENDIF
     \ENDFOR
 \FOR{$i \in [0, 2W)$; $j \in [0,2H)$}
  \IF{ $P_{i, j}^u  == P_{i//2+1, j//2+1}^d$ \AND $M^{correction}_{i, j} != 1$}
    \STATE {$ {F'}_{i, j}^u = F^{d}_{i//2+1, j//2+1} $}
    \STATE $M^{correction}_{i, j} = 1$
  \ENDIF
     \ENDFOR

\end{small}
\end{algorithmic}
\end{algorithm}




To correct the misalignment in the $2\times2$ upsampling layer, we scan the four neighbors of each pixel.
In the $2\times2$ window,
if we find a match between the corresponding pixels' panoptic identity in higher resolution and a neighboring pixels panoptic identity in lower resolution, we copy over that neighboring feature to the corresponding indices in upsampled feature map.
This method is depicted in Algorithm  \ref{alg:misalign}.
Note that the first for loop would correspond to the nearest neighbor upsampling algorithm if there was no if statement in the loop.
We also update the mask, $\mathbf{M^{correction}}$, to keep track of which indices have been successfully aligned.
In the subsequent for loops, for the indices that were not aligned yet, we check if any of the other neighbors are matching the panoptic identity with them.

After Algorithm  \ref{alg:misalign}, we end up with a partially filled upsampled feature map ${\mathbf{F'}^{u}}$ and a $\mathbf{M^{correction}}$ mask which defines which coordinates found a match.
After that we calculate the final $\mathbf{{F'}^{u}}$ by:
\begin{gather*}
{F'}^{u}_{(i,j)}= {F'}^{u}_{(i,j)} + \\ \underbrace{(1-M^{correction}_{i, j}) * f_{holefilling}(S^u_{(i,j)})}_{\text{Hole Filling}}
\end{gather*}

We generate $f_{holefilling}$ with the \textit{Panoptic aware convolution} layer by feeding semantic map  ($\mathbf{S^u}$) as input and panoptic map ($\mathbf{P^u}$) for guidance.
We use the semantic map to encode features from $K\times 2W \times 2H$ semantic map where $K$ is the number of classes to a higher dimension $C \times 2W \times 2H$ with this layer.

\begin{equation}
    f_{holefilling}= PanopticAwareConvolution(S^u)
\end{equation}

With \textit{Panoptic aware upsampling} layer, the features that have been tailored for a specific instance or a semantic map are not copied over to an another one, which improves the accuracy of the generated images.





\begin{figure*}[!ht]
    \centering
    \includegraphics[width=0.9\textwidth]{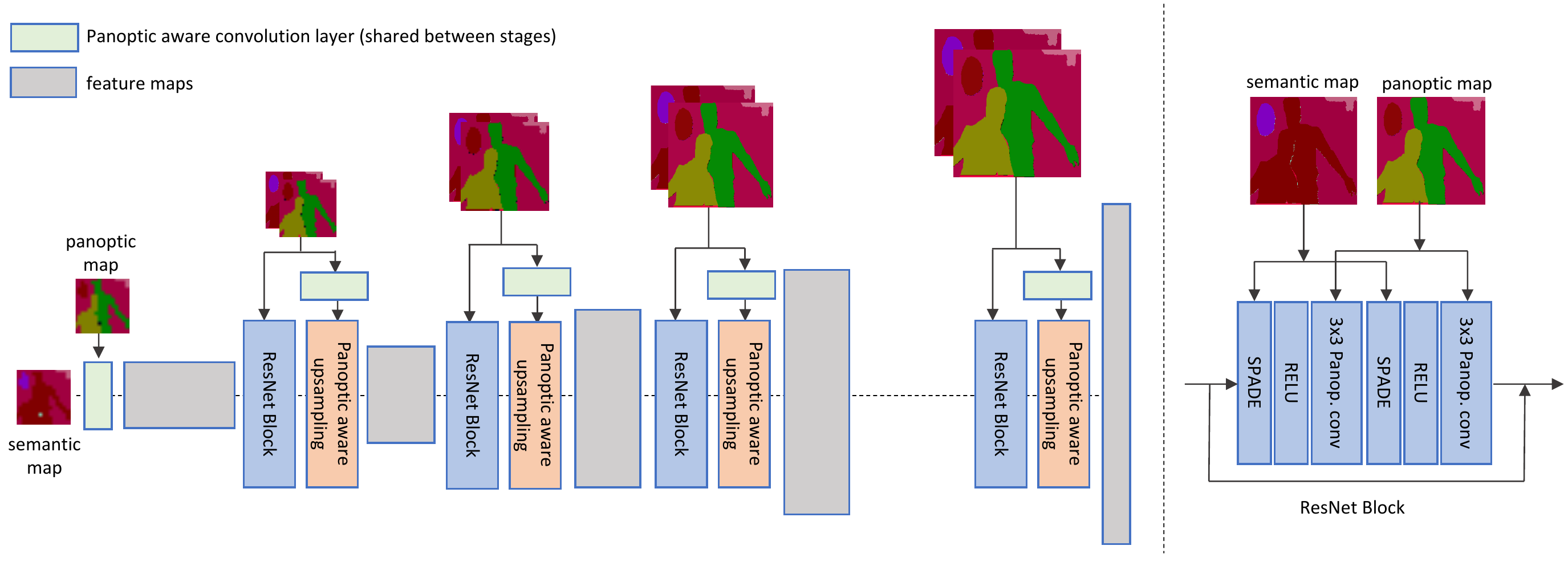}
    \caption{In our generator, each ResNet Block layer uses segmentation and panoptic masks to modulate the layer activations. (Left) The generator contains a series of the residual blocks with Panoptic aware convolution and upsampling layers. (Right) Structure of residual blocks.  }
    \label{fig:arch}
\end{figure*}

\subsection{Network Architecture}
Our final proposed architecture, motivated by SPADE~\cite{park2019semantic} is described in Figure~\ref{fig:arch}. 
Similar to SPADE, we feed a downsampled segmentation map to the first layer of the generator, but in our architecture, it is a Panoptic aware convolution layer that encodes features from $\text{\#Classes}\times W \times H$ semantic map to a higher dimension $1024 \times W \times H$.
In the rest of the network, we replace all convolution layers in the ResNet Blocks with the Panoptic aware convolution layers and all upsampling layers with Panoptic aware upsampling layers.
Each block operates at a different scale and we downsample the semantic and panoptic maps to match the scale of the features.

The input to the SPADE module is kept as a semantic map which learns denormalization parameters. Panoptic maps are not suitable for this computation, since the convolution operation expects a fixed number of channels.
Hence, we rely on SPADE to provide the network with the correct statistics of features based on the semantic classes.

We feed panoptic maps to the panoptic aware convolution layers in order to perform convolution operation based on the instances and classes.
The original full resolution panoptic and semantic maps are also fed to the panoptic aware upsampling layers to perform upsampling alignment correction and hole filling.

The panoptic aware convolution layer that is in the first layer of the architecture which encodes features from $\text{\#Classes}\times W \times H$ semantic map to a higher dimension encoded features are shared between panoptic aware upsampling layers in the rest of the network.
When the number of channels the partial convolution layer from the first layer generates does not match the one expected at different blocks, we decrease the dimension with $1\times1$ convolution layers.
This layer is depicted by green boxes in Figure \ref{fig:arch}. Note that the green box is depicted multiple times in Figure but they are shared between stages.
By sharing the weights, we do not introduce additional parameters to the baseline except the negligible cost of $1\times1$ convolutions. Sharing these weights also makes sense since the task of this layer is common at each stage which is to generate features for instances and semantic classes that appear for the first time at that stage.

\section{Experiments}

\noindent\textbf{Datasets.} We conduct our experiments on Cityscapes \cite{cordts2016cityscapes} and COCO-Stuff \cite{caesar2018coco} datasets that have both instance and semantic segmentation labels available. The Cityscapes dataset contains 3,000 training images and 500 validation images of urban street scenes along with 35 semantic classes and 9 instance classes. All classes are used while synthesizing images but only 19 classes are used for semantic evaluation as defined by Cityscapes evaluation benchmark. COCO-Stuff dataset has 118,000 training images and 5,000 validation images from both indoor and outdoor scenes. This dataset has 182 semantic classes and 81 instance classes. 

\noindent\textbf{Implementation Details.} We use the parameters provided by SPADE baseline \cite{park2019semantic}. Specifically, we use synchronized batch normalization to collect statistics across GPUs, and apply Spectral Norm \cite{miyato2018spectral} to all layers in the generator and discriminator.
We train and generate images in $256\times256$ resolution for COCO-Stuff, and $256\times512$ for Cityscapes dataset.
We train 200 epochs on Cityscapes dataset with batch size 16, and linearly decay the learning rate after 100 epochs as done by \cite{park2019semantic}.
COCO-Stuff dataset is trained for 100 epochs with batch size of 48 with constant learning rate.
Initial learning rates are set to $0.0001$ and $0.0004$ for the generator and discriminator respectively, and networks are trained with ADAM
solver with $\beta_1 = 0$ and $\beta_2 = 0.999$.

\noindent\textbf{Performance Metrics.} We adopt the evaluation metrics as previous conditional image synthesis work~\cite{park2019semantic,wang2018high} plus add another metric for detecting successfully generated object instances. The first two metrics, mean intersection over union (mIoU) and overall pixel accuracy (accuracy), are obtained by inferring a state-of-the-art semantic segmentation model on the synthesized images and comparing how well the predicted segmentation mask matches the ground truth semantic map. Additionally, we use Detection Average Precision (detAP) by using a trained object detection network to evaluate instance detection accuracy on synthesized images.

We  use the same segmentation networks used in ~\cite{park2019semantic} for evaluation. Specifically, we use DeepLabV2~\cite{chen2018deeplab,kazuto2018} for COCO-Stuff, and DRN-D-105~\cite{yu2017dilated} for Cityscapes datasets.
For detection, we use Faster-RCNN \cite{ren2015faster} with ResNet-50 backbone.
In addition to the mIoU, accuracy and  detAP  performance metrics, we use the Fr\'echet Inception Distance (FID)~\cite{heusel2017gans} to measure the distance between the distribution of synthesized results and the distribution of real images.

\noindent\textbf{Baselines.} We compare our method against three popular image synthesis frameworks, namely: cascaded refinement network (CRN) \cite{chen2017photographic}, semi-parametric image synthesis (SIMS) \cite{qi2018semi}, and  spatially-adaptive denormalization model (SPADE) \cite{park2019semantic}. CRN uses a deep network with given semantic label map, it repeatedly refines the output from low to high resolution without an adversarial training.
SIMS uses a memory bank of image segments constructed from a training set of images and refines the boundaries via a deep network. Both SIMS and CRN operate only on a semantic map.
SPADE is the current state-of-the-art conditional image synthesis method, and not only uses semantic map but also incorporates the instance information via a boundary map. The pixel in the boundary map is 1 if its object identity is different from any of its 4 neighbors, and 0 otherwise. This approach does not provide the full instance information especially in cluttered scenes with lots of objects occluding each other.
We compare with SIMS on Cityscapes dataset but not on COCO-stuff dataset as SIMS requires queries to the training set images and it is computationally costly for a large dataset such as the COCO-stuff dataset. 

\begin{table}
    \centering
    \begin{tabular}{l|c|c|c|c}
        Method &  detAP & mIoU & accuracy &  FID\\
        \hline
        CRN \cite{chen2017photographic}&  8.75 & 52.4 & 77.1 &   104.7 \\
        SIMS \cite{qi2018semi} &  2.60 & 47.2 & 75.5 &  \textbf{49.7} \\
        SPADE \cite{park2019semantic} & 11.67 & 62.3 & 81.9 &  71.8\\
        SPADE* & 11.80 & 62.2 & 81.9 &  94.0\\
        Ours & \textbf{13.43} & \textbf{64.8} & \textbf{82.4} &  96.4\\
        
    \end{tabular}
    \caption{Results on Cityscapes. Our method outperforms current leading methods in detAP, mIoU and overall pixel accuracy. SPADE* is trained by us.}
    \label{tab:cityscapes}
\end{table}

\begin{table}
    \centering
    \begin{tabular}{l|c|c|c|c}
        Method & detAP &  mIoU & accuracy  & FID\\
        \hline
        CRN \cite{chen2017photographic} &  22.7 & 23.7  & 40.4  & 70.4 \\
        SPADE \cite{park2019semantic} & 28.5 & 37.4 & 67.9 &  \textbf{22.6}\\
        SPADE* & 29.0 & 38.2 & 68.6 &  25.3\\
        Ours & \textbf{31.0} & \textbf{38.6} & \textbf{69.0} &  28.8\\
        
    \end{tabular}
    \caption{Results on COCO-Stuff. Our  method  outperforms  current leading methods in detAP, mIoU and overall pixel accuracy. SPADE* is trained by us.}
    \label{tab:coco}
\end{table}

\newcommand{\interpfigk}[1]{\includegraphics[width=3.4cm,height=2.0cm]{#1}}

\begin{table*}[t]
    \label{tab:city_classes}
	\begin{center}
		\resizebox{2.1\columnwidth}{!}{
		\begin{tabular}{  c |  c  c  c  c  c c  c  c c  c  c c  c  c c  c  c c  c }
		\hline
		Method	&  road  & swalk & build. & wall & fence & pole & tlight & tsign & veg. & terr. & sky & person & rider & car & truck & bus & train & mbike &  bike   \\
		\hline
		\hline
		CRN \cite{chen2017photographic}   &96.9& 79.5& 76.7& 29.0& 10.6& 34.8& 39.8& 44.3& 68.4& 54.4& 91.9& 63.0& 39.7& 87.8& 25.0& 56.2& 31.8& 14.5& 52.2  \\
		SIMS \cite{qi2018semi} &  93.3 & 66.1 & 73.6 & 33.1&  34.5&  30.3&  27.2&  39.5& 73.4& 46.2&  56.6&  42.9&  31.0&  70.3&  35.8&  42.5&  37.3&  20.3&  43.1  \\
		SPADE \cite{park2019semantic} & 97.4& 80.0& 87.9& 50.6& 47.2& \textbf{35.9}& 39.0& 44.7& 88.2& 66.1& 91.6& 62.3& \textbf{38.7}& 88.7& 65.0 & 70.2& 41.4& 28.6& 58.8   \\
		
		Ours & \textbf{97.7} & \textbf{82.5} & \textbf{89.2}&  \textbf{60.6}& \textbf{54.2}& 35.3 & \textbf{39.8} & \textbf{50.0}& \textbf{89.5}& \textbf{69.0}& \textbf{92.4} & \textbf{63.2} & 38.2 & \textbf{90.6} & \textbf{66.7} & \textbf{72.2} & \textbf{48.8} & \textbf{31.2}& \textbf{59.1}\\	
	  \end{tabular}
	}
	\end{center}
	\caption{Per-class mIoU results on Cityscapes.}
	\label{table:class_miou}
\end{table*}

\begin{figure*}[]
  \centering
  \setlength\tabcolsep{0.32pt}
  \renewcommand{\arraystretch}{0.15}
  \begin{tabular}{cccccc}
    Panoptic Map &CRN \cite{chen2017photographic} &SIMS \cite{qi2018semi} & SPADE \cite{park2019semantic} &Ours\\
    \interpfigk{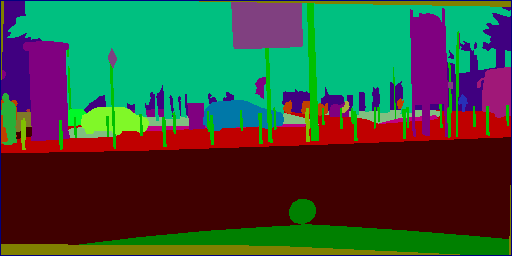}&
    \interpfigk{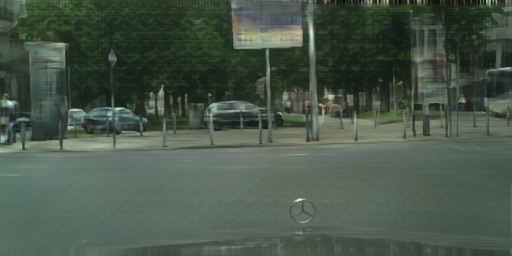}&
     \interpfigk{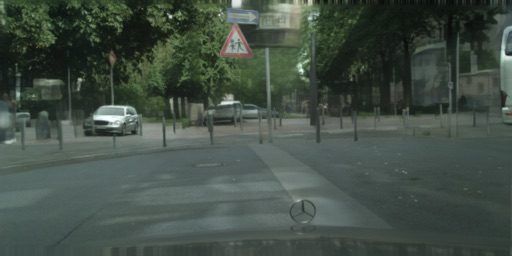}&
     \interpfigk{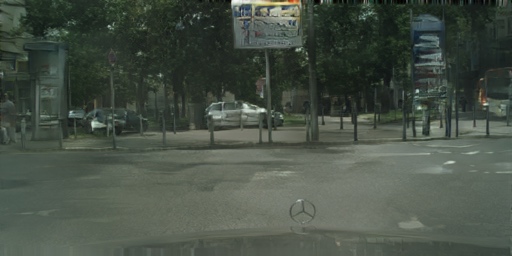}&
     \interpfigk{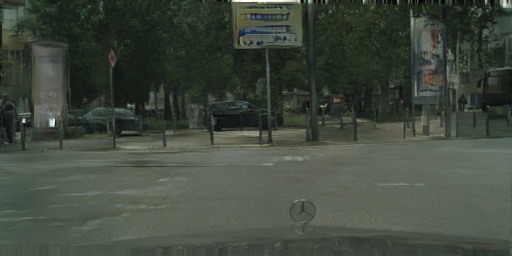}\\
     &
    \interpfigk{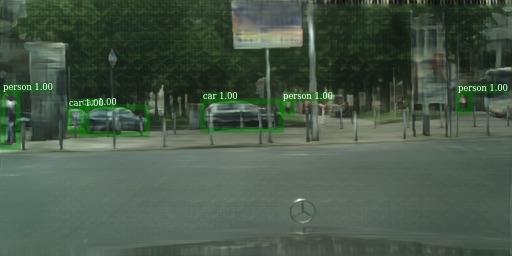}&
     \interpfigk{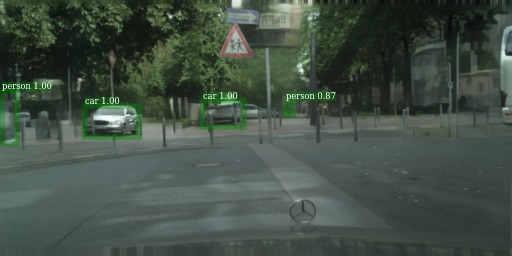}&
     \interpfigk{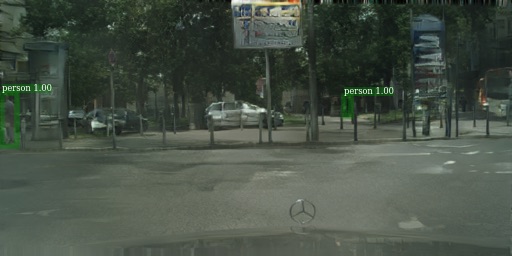}&
     \interpfigk{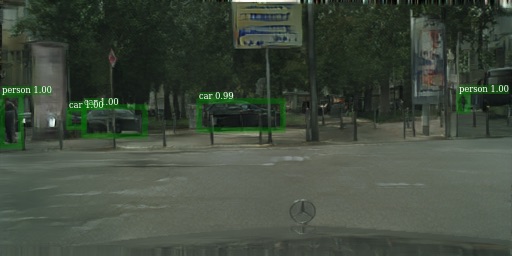}\\
    \interpfigk{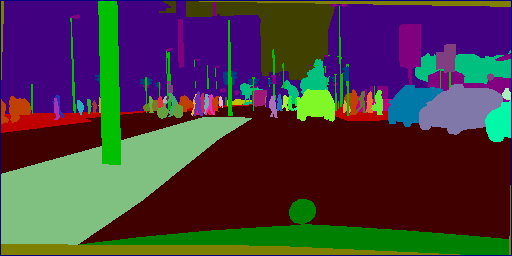}&
    \interpfigk{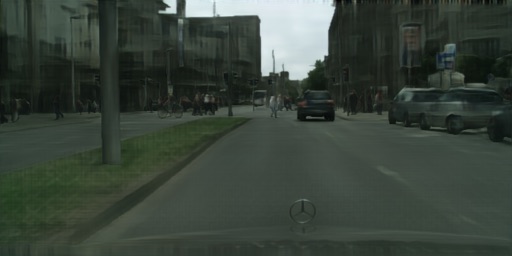}&
     \interpfigk{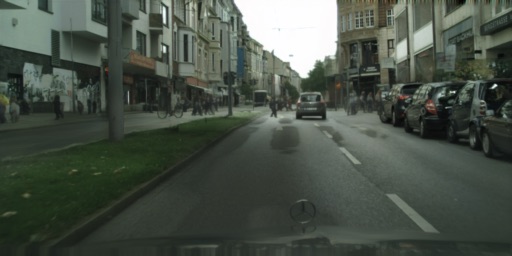}&
     \interpfigk{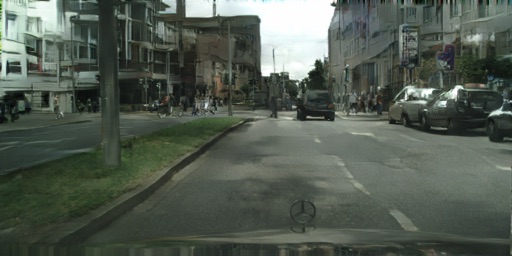}&
     \interpfigk{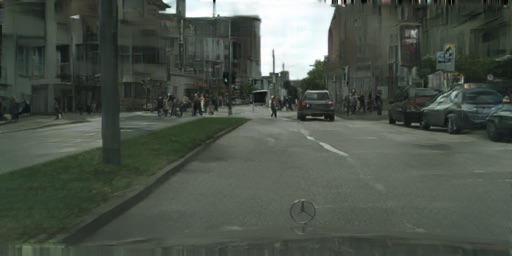}\\
     &
    \interpfigk{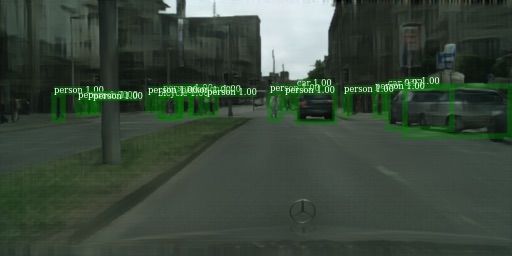}&
     \interpfigk{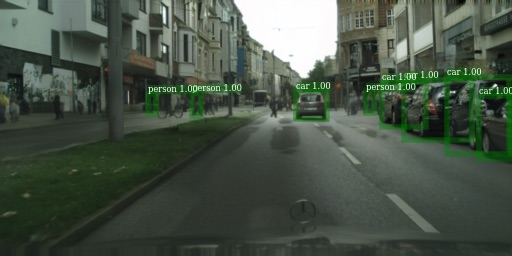}&
     \interpfigk{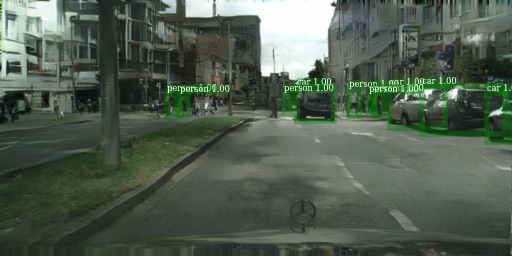}&
     \interpfigk{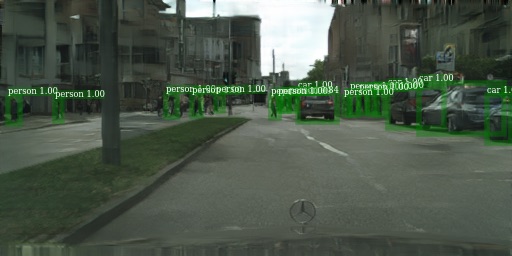}\\
     \end{tabular}
     \caption{Visual comparison of image synthesis results on Cityscapes dataset. We also provide the bounding box detection predictions from Faster-RCNN. The cars in the first row images are occluded by poles which create a challenge for the image synthesis methods. CRN generates cars that can be detected by Faster-RCNN but visually look less pleasing. SIMS loosely follows the provided semantic map, and the cars SPADE generates are not distinctive enough to be detected by Faster-RCNN. In the third row, cars generated on the right side of the images present a challenge for the algorithms that only use semantic maps as seen in the images from CRN and SIMS where CRN generates two cars and SIMS generates four cars while three cars should be present. Thanks to the boundary maps used in SPADE, it can generate the correct number of cars. However, our proposed method along with generating the correct number of car instances, also generates more instances of persons that can be detected with higher accuracy.}
     \label{fig:city_qualitative}
\end{figure*}

\noindent\textbf{Quantitative Results.} In Tables \ref{tab:cityscapes} and \ref{tab:coco}, we provide the results for Cityscapes and COCO-Stuff datasets, respectively.
We find that our method outperforms the current state-of-the-art by a large margin for object detection score, mIoU and pixel level accuracy in both datasets.
Table \ref{tab:city_classes} reports mIoU for each class in Cityscapes dataset.
We improve almost all of the classes significantly.
Especially, our proposed method improves mIoU for traffic sign from $44.7$ to $50.0$, which is a challenging class because of the small size of the signs.

We observe a slight degradation in our FID score  compared to the released SPADE models, and the SPADE models we trained with the parameters provided by \cite{park2019semantic}.
FID score tries to match the variances/diversities between real and generated images, without caring about the correspondence with the conditioned semantic map and instance map. Our results have better correspondence with the underlying semantic and instance maps.
Though this would be the desired behavior, the results may get affected by human annotation bias.
We suspect, such annotation bias (e.g. straight line bias, over-simplified polygonal shape bias) in the inputs may deteriorate the matching of variances.
Also note that SIMS produces images that have significantly lower FID score than the other methods even though it achieves worse detAP and mIoU scores.
This is because SIMS copies image patches from training dataset, and sometimes the copied patch does not faithfully match the given segmentation mask. This issue  becomes even more apparent in detAP score, as SIMS copies over patches without ensuring the number of cars being consistent with the panoptic map.

\noindent\textbf{Qualitative Comparison.} In Figures \ref{fig:city_qualitative} and \ref{fig:coco_qualitative}, we provide image synthesis results of our method and other competing methods. We also provide the bounding box detection predictions from Faster-RCNN.
We especially provide examples where multiple instances occlude each other.
We find that our method produces instances with better visual quality in challenging scenarios.
Specifically, we find that our method generates distinct cars even when they are behind poles, and can generate detectable people even when they are far away as shown in Figure \ref{fig:city_qualitative}.
As can be seen in Figure \ref{fig:coco_qualitative}, we find that other methods may blend the pattern and texture of objects among neighboring instances, whereas our method clearly separates them.

\begin{table}[]
    \centering
    \begin{tabular}{l|c|c}
        Method &  mIoU & detAP\\
        \hline
        Baseline (SPADE) & 60.00 & 10.97 \\
        +Panoptic-aware Partial Conv & 61.24 & 11.50\\
        +Panoptic-aware Upsampling & 64.55 & 13.04\\
    \end{tabular}
    \caption{Ablation studies on Cityscapes dataset. Results are averaged over 3 runs, and they are slightly different than the results in Table \ref{tab:cityscapes}.}
    \label{tab:ablation}
\end{table}

\newcommand{\interpfigc}[1]{\includegraphics[width=4cm,height=4cm]{#1}}
\newcommand{\interpfigd}[1]{\includegraphics[width=4cm,height=4cm]{#1}}
\begin{figure*}[]
  \centering
  \setlength\tabcolsep{0.32pt}
  \renewcommand{\arraystretch}{0.15}
  \begin{tabular}{ccccc}
    Panoptic Map &CRN \cite{chen2017photographic}  & SPADE \cite{park2019semantic} &Ours\\
    \interpfigc{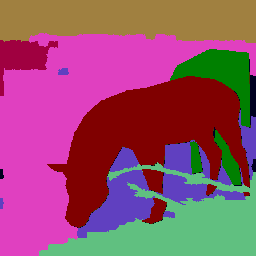}&
    \interpfigd{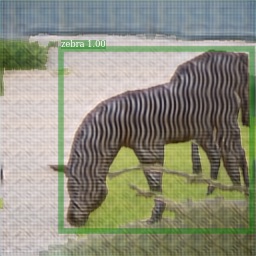}&
    \interpfigd{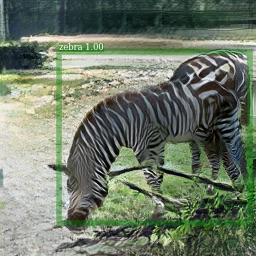}&
    \interpfigd{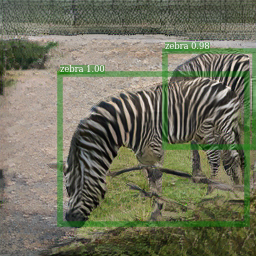}&
    \\
    \interpfigc{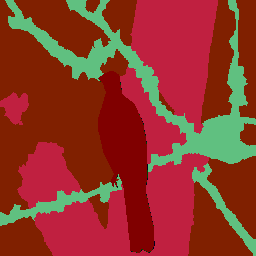}&
    \interpfigc{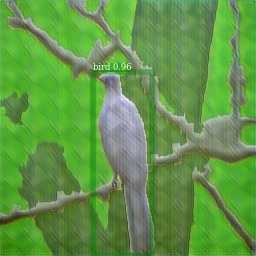}&
    \interpfigc{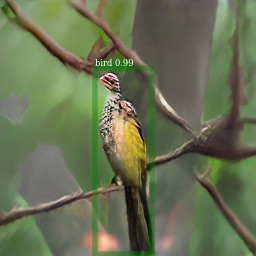}&
    \interpfigc{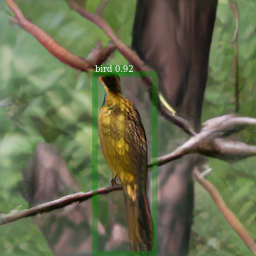}&
    \\
    \interpfigc{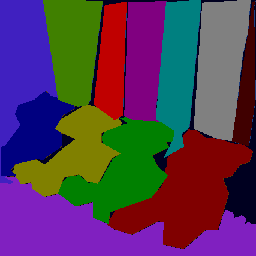}&
    \interpfigc{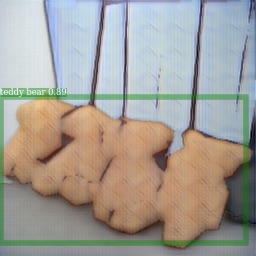}&
    \interpfigc{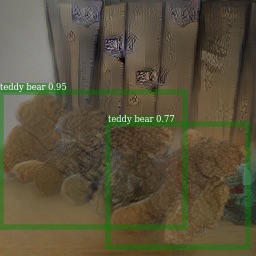}&
    \interpfigc{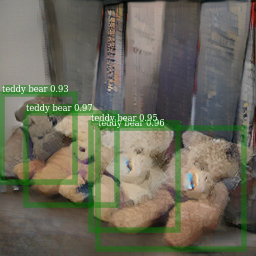}&
    \\
    \interpfigc{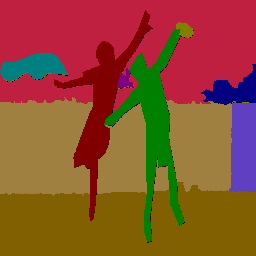}&
    \interpfigc{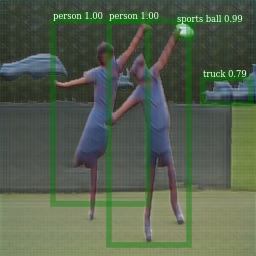}&
    \interpfigc{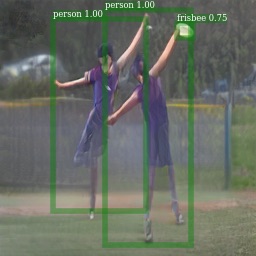}&
    \interpfigc{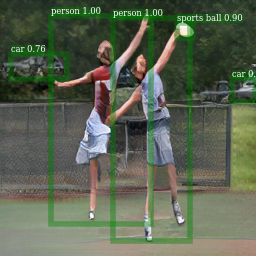}&
     \end{tabular}
     \caption{Visual comparison of  image synthesis results on the COCO-stuff dataset. We also display the bounding box detection predictions from Faster-RCNN. Other methods generate patterns that are continues throughout instances which makes the instances indistinguishable. Also note that in the last row our method is able to produce detectable car instances in a cluttered scene.  }
     \label{fig:coco_qualitative}
\end{figure*}

\noindent\textbf{Ablation Studies.} We conduct controlled experiments and gradually add our proposed components.
We start with a baseline SPADE model \cite{park2019semantic}.
We train the model three times, and report the average results. First, we replace the convolutions in ResNet blocks and the first layer with panoptic aware convolution layers.
Second, we additionally replace nearest neighbor upsampling layers with panoptic aware upsampling layers.
The segmentation mIoU scores and detAP scores of
the generated images by each setup are shown in Table \ref{tab:ablation} where each added module increases the performance.


\section{Conclusion}

In conclusion, we propose a panoptic-based image synthesis network that generates images with higher fidelity to the underlying segmentation and instance information.
We show that our method is better at generating distinct instances in challenging scenarios and outperforms previous state-of-the-art significantly in detAP metric, a metric which has not been used to evaluate conditional image synthesis results before.

\noindent\textbf{Future Work.}
Multi-modal image synthesis and controllability of styles are very important for content generation applications.
The architecture in our experiments does not support style-guided image synthesis.
However, our work can be extended to output multiple styles via an encoder-decoder architecture as proposed in pix2pixHD \cite{wang2018high}.
Furthermore, the proposed panoptic aware convolution and upsampling layers can be used for feature maps that decode styles, and can provide further improvements.
We leave this as future work.


{\small
\bibliographystyle{ieee_fullname}
\bibliography{egbib}
}

\begin{figure*}[]
  \centering
   \caption{Additional results on the COCO-Stuff dataset.}
  \setlength\tabcolsep{0.36pt}
  \renewcommand{\arraystretch}{0.15}
  \begin{tabular}{ccccc}
   \rotatebox{90}{~~~~~Panoptic Map \& GT} &
    \interpfigc{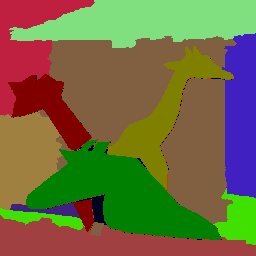} &
    \interpfigc{} &
    \interpfigc{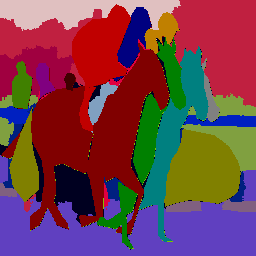} &
    \interpfigc{}  \\
     \rotatebox{90}{~~~~~~~~~~~~~~~Ours} &
    \interpfigc{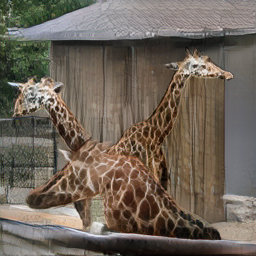} & 
    \interpfigc{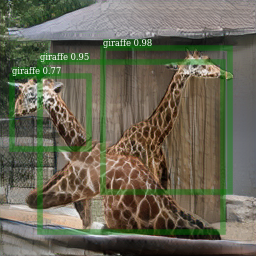}&
    \interpfigc{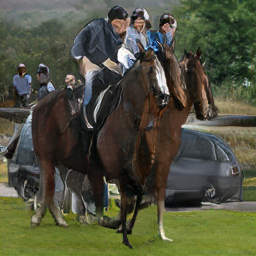} & 
    \interpfigc{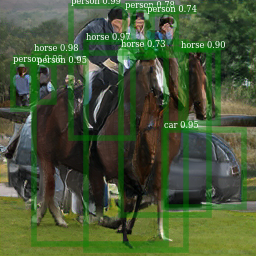} \\
     \rotatebox{90}{~~~~~~~~~~~~~~~SPADE \cite{park2019semantic}} &
    \interpfigc{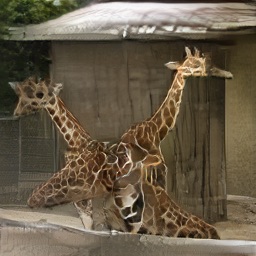}&
     \interpfigc{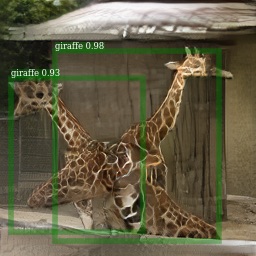}&
    \interpfigc{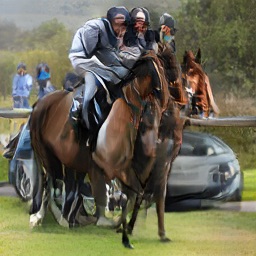}&
     \interpfigc{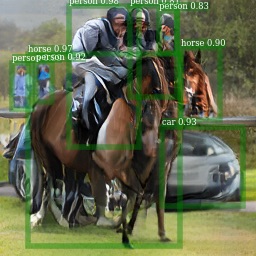}\\
    \rotatebox{90}{~~~~~~~~~~~~~~~CRN \cite{chen2017photographic}}&
    \interpfigc{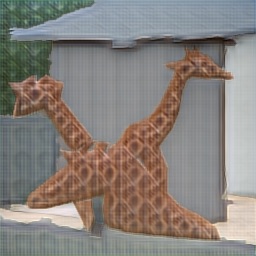}&
    \interpfigc{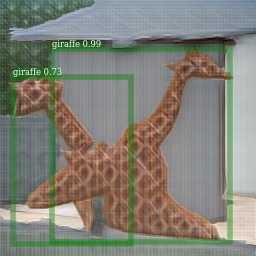} &
    \interpfigc{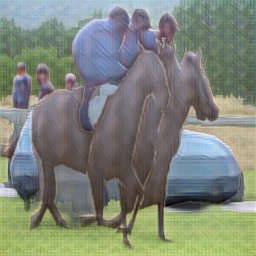}&
    \interpfigc{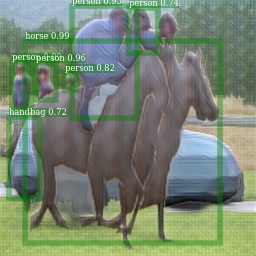}\\
     \end{tabular}
     \label{fig:coco_qualitative_1}
\end{figure*}

\begin{figure*}[]
  \centering
  \caption{Additional results on the COCO-Stuff dataset.}
  \setlength\tabcolsep{0.36pt}
  \renewcommand{\arraystretch}{0.15}
  \begin{tabular}{ccccc}
   \rotatebox{90}{~~~~~Panoptic Map \& GT} &
    \interpfigc{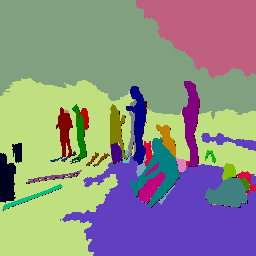} &
    \interpfigc{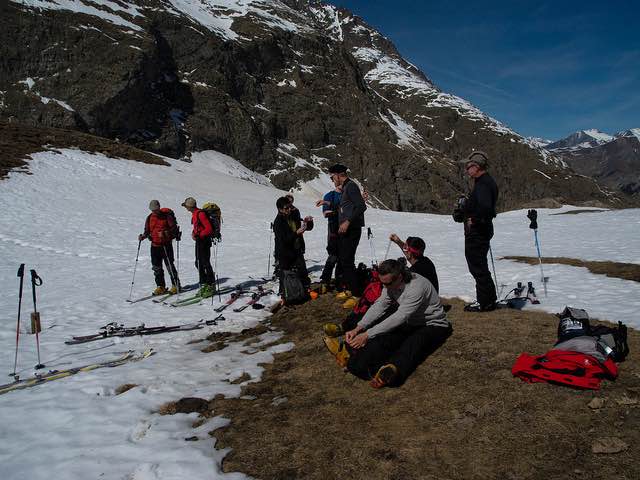} &
    \interpfigc{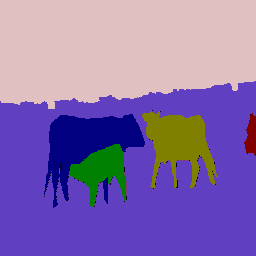} &
    \interpfigc{}  \\
     \rotatebox{90}{~~~~~~~~~~~~~~~Ours} &
    \interpfigc{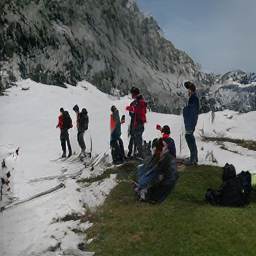} & 
    \interpfigc{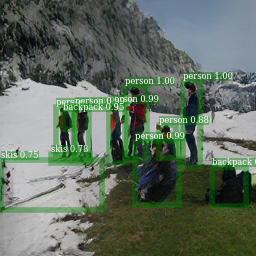}&
    \interpfigc{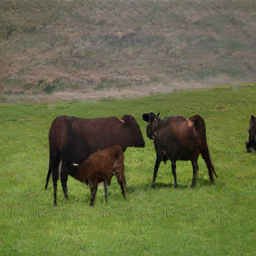} & 
    \interpfigc{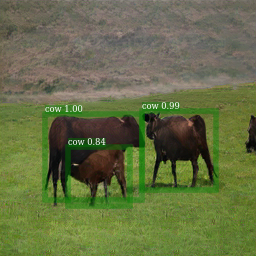} \\
     \rotatebox{90}{~~~~~~~~~~~~~~~SPADE \cite{park2019semantic}} &
    \interpfigc{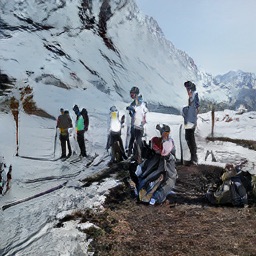}&
     \interpfigc{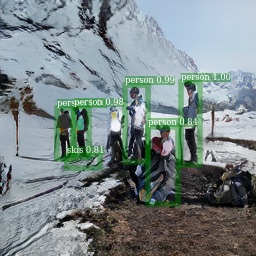}&
    \interpfigc{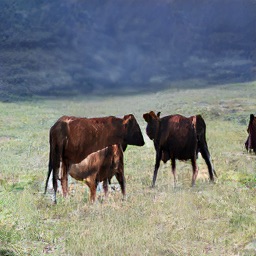}&
     \interpfigc{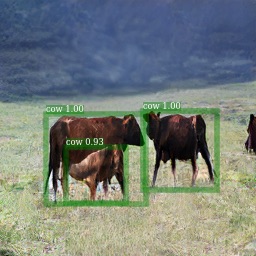}\\
    \rotatebox{90}{~~~~~~~~~~~~~~~CRN \cite{chen2017photographic}}&
    \interpfigc{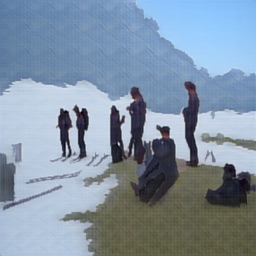}&
    \interpfigc{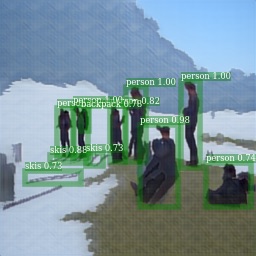} &
    \interpfigc{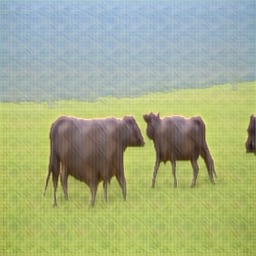}&
    \interpfigc{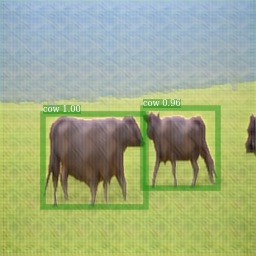}\\
     \end{tabular}
     \label{fig:coco_qualitative_2}
\end{figure*}

\begin{figure*}[]
  \centering
\caption{Additional results on the COCO-Stuff dataset.}
  \setlength\tabcolsep{0.36pt}
  \renewcommand{\arraystretch}{0.15}
  \begin{tabular}{ccccc}
   \rotatebox{90}{~~~~~Panoptic Map \& GT} &
    \interpfigc{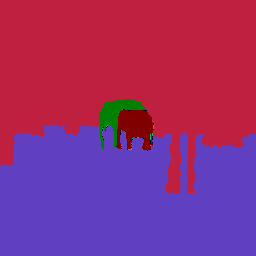} &
    \interpfigc{} &
    \interpfigc{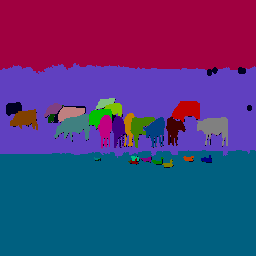} &
    \interpfigc{}  \\
     \rotatebox{90}{~~~~~~~~~~~~~~~Ours} &
    \interpfigc{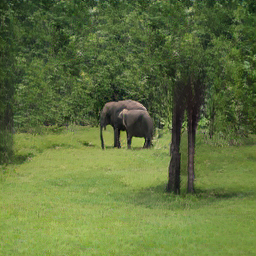} & 
    \interpfigc{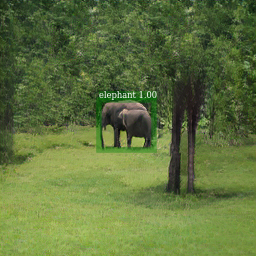}&
    \interpfigc{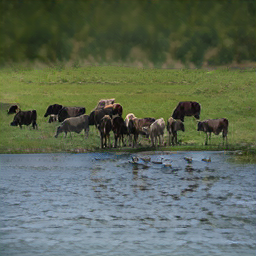} & 
    \interpfigc{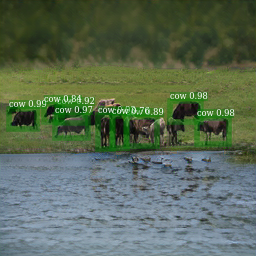} \\
     \rotatebox{90}{~~~~~~~~~~~~~~~SPADE \cite{park2019semantic}} &
    \interpfigc{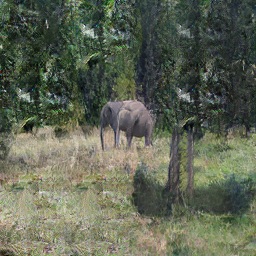}&
     \interpfigc{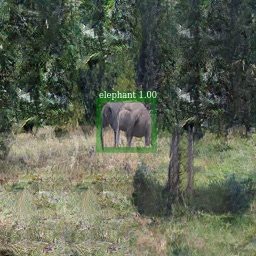}&
    \interpfigc{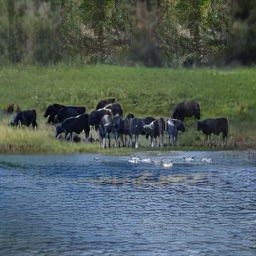}&
     \interpfigc{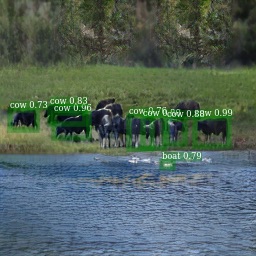}\\
    \rotatebox{90}{~~~~~~~~~~~~~~~CRN \cite{chen2017photographic}}&
    \interpfigc{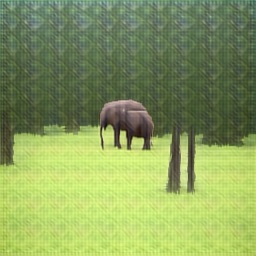}&
    \interpfigc{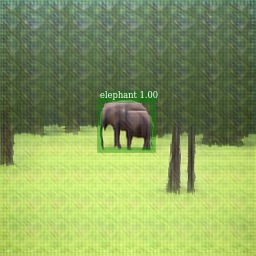} &
    \interpfigc{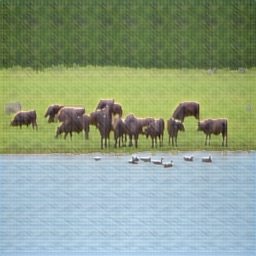}&
    \interpfigc{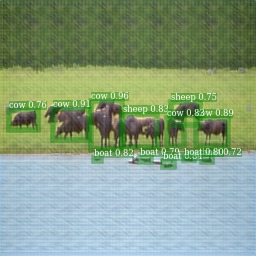}\\
     \end{tabular}
     \label{fig:coco_qualitative_3}
\end{figure*}

\newcommand{\interpfigcs}[1]{\includegraphics[width=8cm,height=4cm]{#1}}
\begin{figure*}[]
     \caption{Additional results on the Cityscapes dataset.}
  \centering
  \setlength\tabcolsep{0.36pt}
  \renewcommand{\arraystretch}{0.15}
  \begin{tabular}{ccc}
     \rotatebox{90}{~~~~~Panoptic Map \& GT} &
    \interpfigcs{Figures/cityscapes/label_ins/frankfurt_000000_011461_leftImg8bit.png} & 
    \interpfigcs{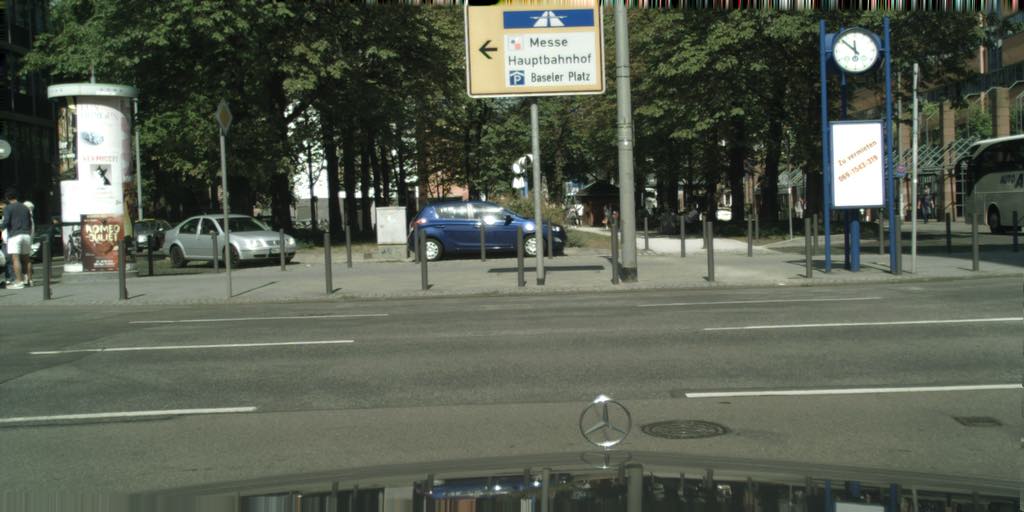} \\
    \rotatebox{90}{~~~~~~~~~~~~~~~Ours} &
    \interpfigcs{Figures/cityscapes/ours_jpeg/frankfurt_000000_011461_leftImg8bit.jpg} & 
    \interpfigcs{Figures/cityscapes/ours_det_jpeg/frankfurt_000000_011461_leftImg8bit.jpg} \\
    \rotatebox{90}{~~~~~~~~~~~~~~~SPADE \cite{park2019semantic}} &
    \interpfigcs{Figures/cityscapes/spade_jpeg/frankfurt_000000_011461_leftImg8bit.jpg}&
     \interpfigcs{Figures/cityscapes/spade_det_jpeg/frankfurt_000000_011461_leftImg8bit.jpg}\\
    \rotatebox{90}{~~~~~~~~~~~~~~~CRN \cite{chen2017photographic}}&
    \interpfigcs{Figures/cityscapes/crn_jpeg/frankfurt_000000_011461_leftImg8bit.jpg}&
    \interpfigcs{Figures/cityscapes/crn_det_jpeg/frankfurt_000000_011461_leftImg8bit.jpg}\\
    \rotatebox{90}{~~~~~~~~~~~~~~~SIMS \cite{qi2018semi}}&
    \interpfigcs{Figures/cityscapes/sims_jpeg/frankfurt_000000_011461_leftImg8bit.jpg}&
     \interpfigcs{Figures/cityscapes/sims_det_jpeg/frankfurt_000000_011461_leftImg8bit.jpg}\\

     \end{tabular}

     \label{fig:city_qualitative_1}
\end{figure*}

\begin{figure*}[]
  \centering
 \caption{Additional results on the Cityscapes dataset. }
  \setlength\tabcolsep{0.36pt}
  \renewcommand{\arraystretch}{0.15}
  \begin{tabular}{ccc}
     \rotatebox{90}{~~~~~Panoptic Map \& GT} &
    \interpfigcs{Figures/cityscapes/label_ins/frankfurt_000001_002646_leftImg8bit.png} &
    \interpfigcs{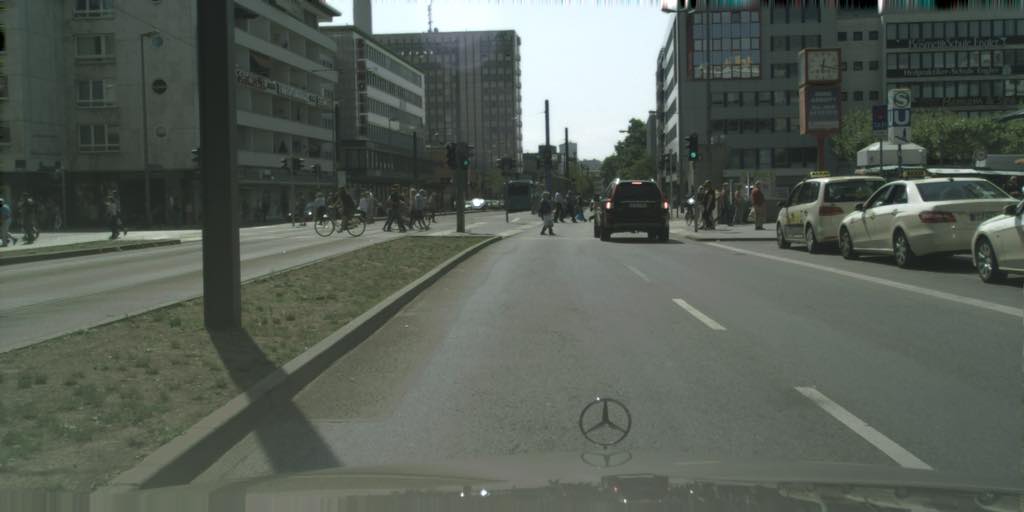} \\
    \rotatebox{90}{~~~~~~~~~~~~~~~Ours} &
    \interpfigcs{Figures/cityscapes/ours_jpeg/frankfurt_000001_002646_leftImg8bit.jpg} & 
    \interpfigcs{Figures/cityscapes/ours_det_jpeg/frankfurt_000001_002646_leftImg8bit.jpg} \\
     \rotatebox{90}{~~~~~~~~~~~~~~~SPADE \cite{park2019semantic}} &
    \interpfigcs{Figures/cityscapes/spade_jpeg/frankfurt_000001_002646_leftImg8bit.jpg}&
     \interpfigcs{Figures/cityscapes/spade_det_jpeg/frankfurt_000001_002646_leftImg8bit.jpg}\\
    \rotatebox{90}{~~~~~~~~~~~~~~~CRN \cite{chen2017photographic}}&
    \interpfigcs{Figures/cityscapes/crn_jpeg/frankfurt_000001_002646_leftImg8bit.jpg}&
    \interpfigcs{Figures/cityscapes/crn_det_jpeg/frankfurt_000001_002646_leftImg8bit.jpg}\\
    \rotatebox{90}{~~~~~~~~~~~~~~~SIMS \cite{qi2018semi}}&
    \interpfigcs{Figures/cityscapes/sims_jpeg/frankfurt_000001_002646_leftImg8bit.jpg}&
     \interpfigcs{Figures/cityscapes/sims_det_jpeg/frankfurt_000001_002646_leftImg8bit.jpg}\\
     \end{tabular}

     \label{fig:city_qualitative_2}
\end{figure*}

\begin{figure*}[]
  \centering
  \caption{Additional results on the Cityscapes dataset.}
  \setlength\tabcolsep{0.36pt}
  \renewcommand{\arraystretch}{0.15}
  \begin{tabular}{ccc}
     \rotatebox{90}{~~~~~Panoptic Map \& GT} &
    \interpfigcs{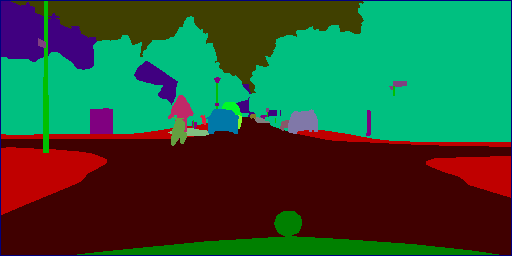} &
    \interpfigcs{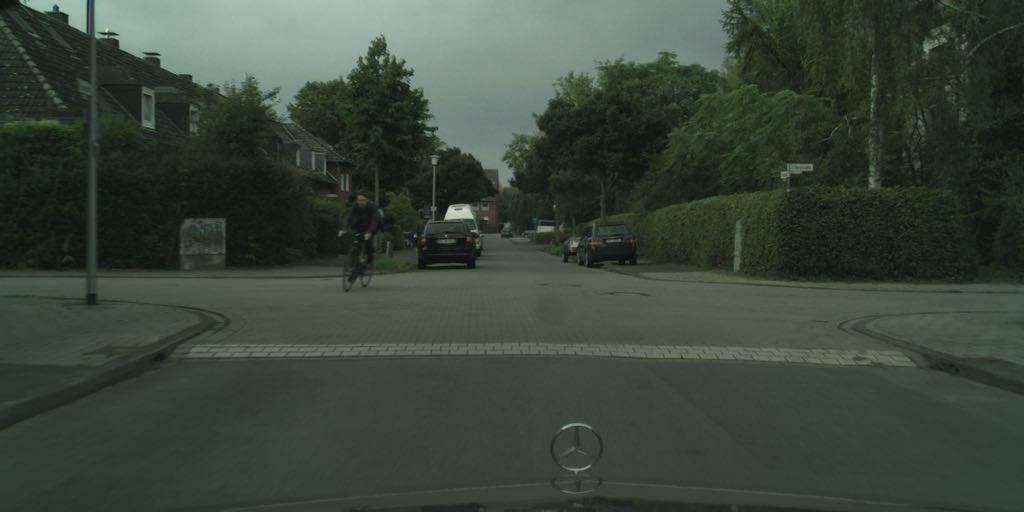} \\
    \rotatebox{90}{~~~~~~~~~~~~~~~Ours} &
    \interpfigcs{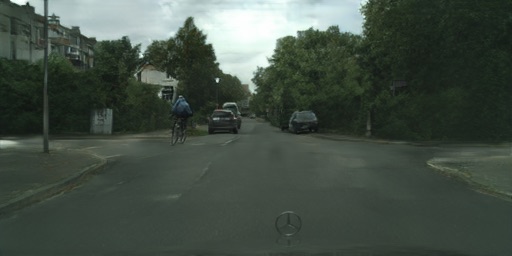} & 
    \interpfigcs{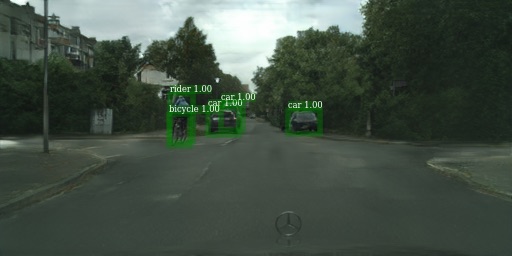} \\
     \rotatebox{90}{~~~~~~~~~~~~~~~SPADE \cite{park2019semantic}} &
    \interpfigcs{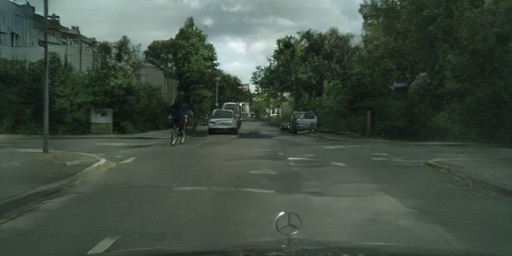}&
     \interpfigcs{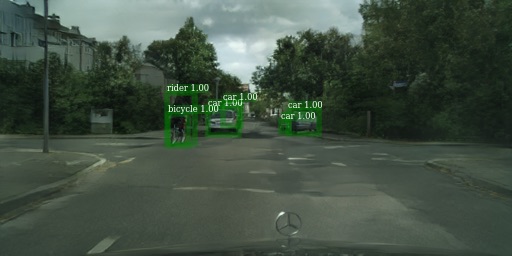}\\
    \rotatebox{90}{~~~~~~~~~~~~~~~CRN \cite{chen2017photographic}}&
    \interpfigcs{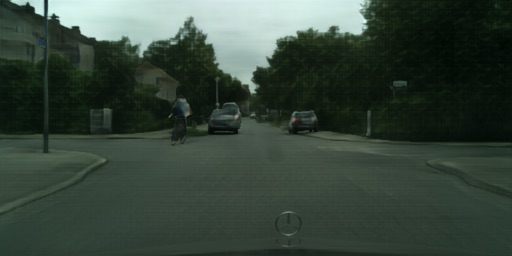}&
    \interpfigcs{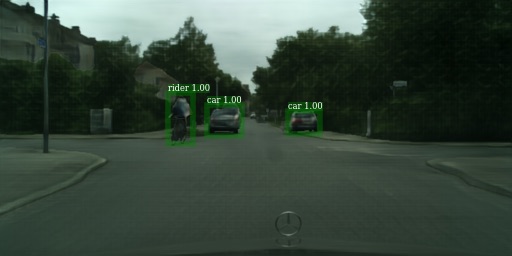}\\
    \rotatebox{90}{~~~~~~~~~~~~~~~SIMS \cite{qi2018semi}}&
    \interpfigcs{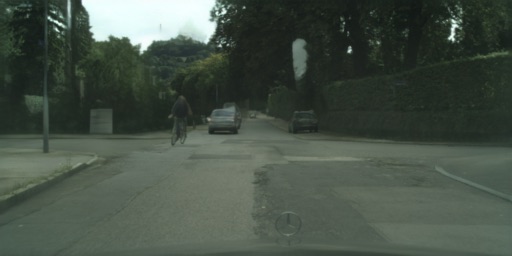}&
     \interpfigcs{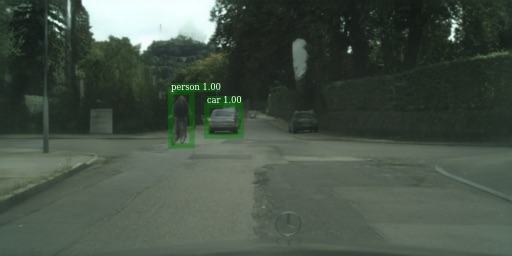}\\
     \end{tabular}
     \label{fig:city_qualitative_3}
\end{figure*}

\begin{figure*}[]
  \centering
     \caption{Additional results on the Cityscapes dataset.}
  \setlength\tabcolsep{0.36pt}
  \renewcommand{\arraystretch}{0.15}
  \begin{tabular}{ccc}
     \rotatebox{90}{~~~~~Panoptic Map \& GT} &
    \interpfigcs{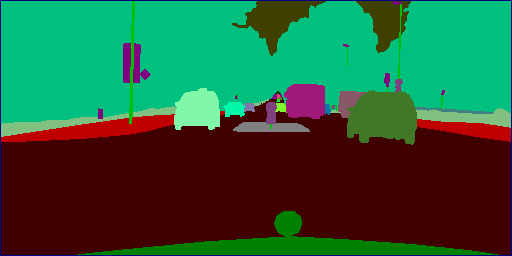} &
    \interpfigcs{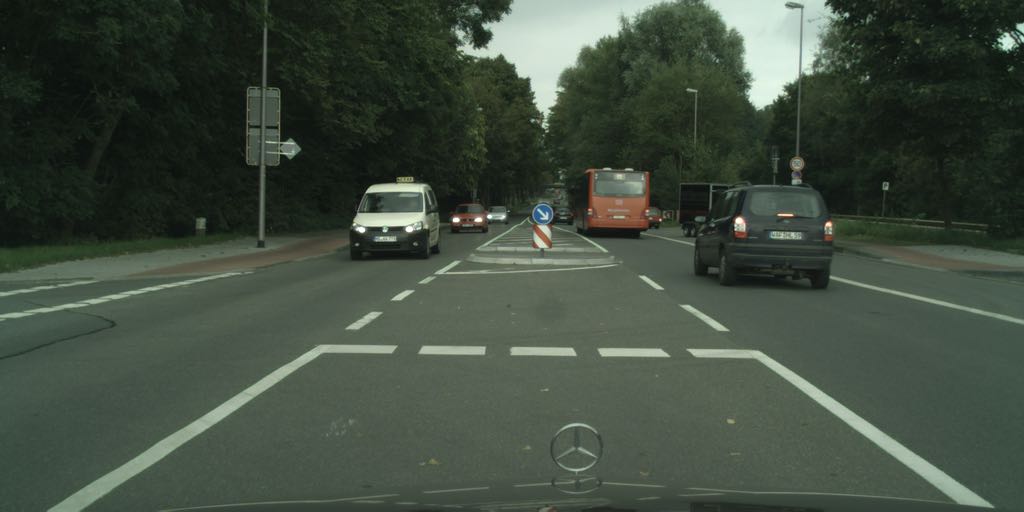} \\
    \rotatebox{90}{~~~~~~~~~~~~~~~Ours} &
    \interpfigcs{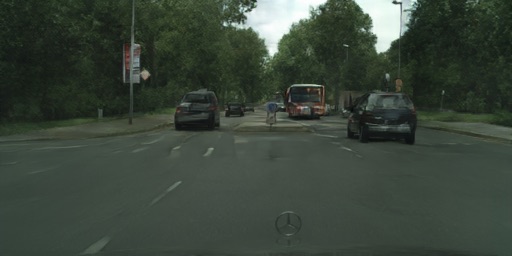} & 
    \interpfigcs{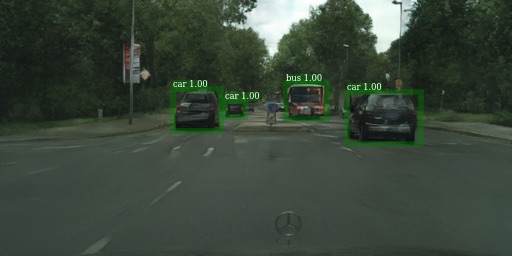} \\
     \rotatebox{90}{~~~~~~~~~~~~~~~SPADE \cite{park2019semantic}} &
    \interpfigcs{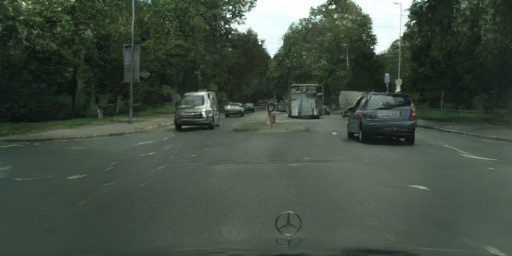}&
     \interpfigcs{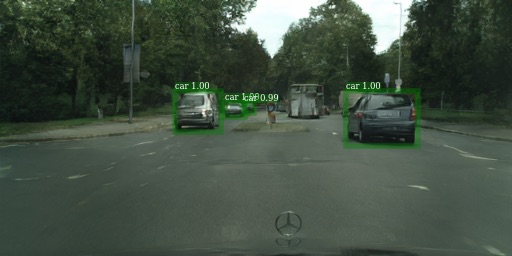}\\
    \rotatebox{90}{~~~~~~~~~~~~~~~CRN \cite{chen2017photographic}}&
    \interpfigcs{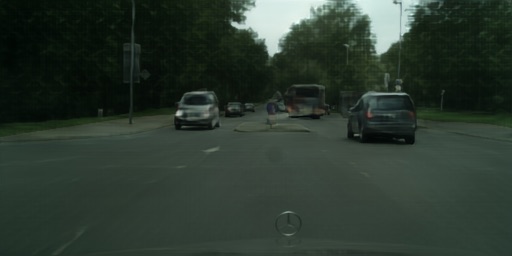}&
    \interpfigcs{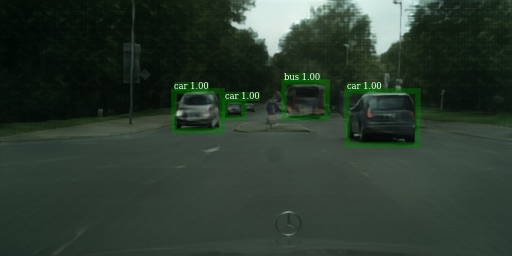}\\
    \rotatebox{90}{~~~~~~~~~~~~~~~SIMS \cite{qi2018semi}}&
    \interpfigcs{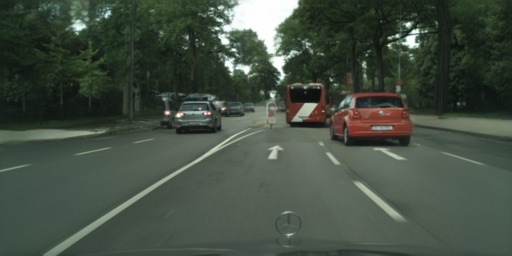}&
     \interpfigcs{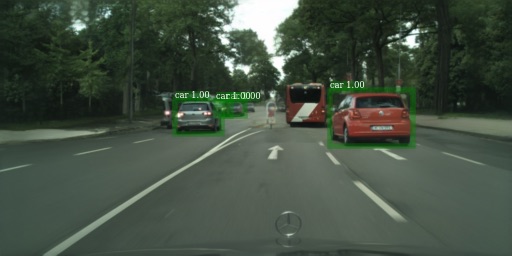}\\
     \end{tabular}
     \label{fig:city_qualitative_4}
\end{figure*}

\begin{figure*}[]
  \centering
  \caption{Additional results on the Cityscapes dataset.}
  \setlength\tabcolsep{0.36pt}
  \renewcommand{\arraystretch}{0.15}
  \begin{tabular}{ccc}
     \rotatebox{90}{~~~~~Panoptic Map \& GT} &
    \interpfigcs{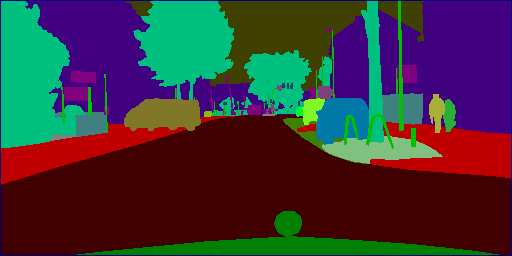} &
    \interpfigcs{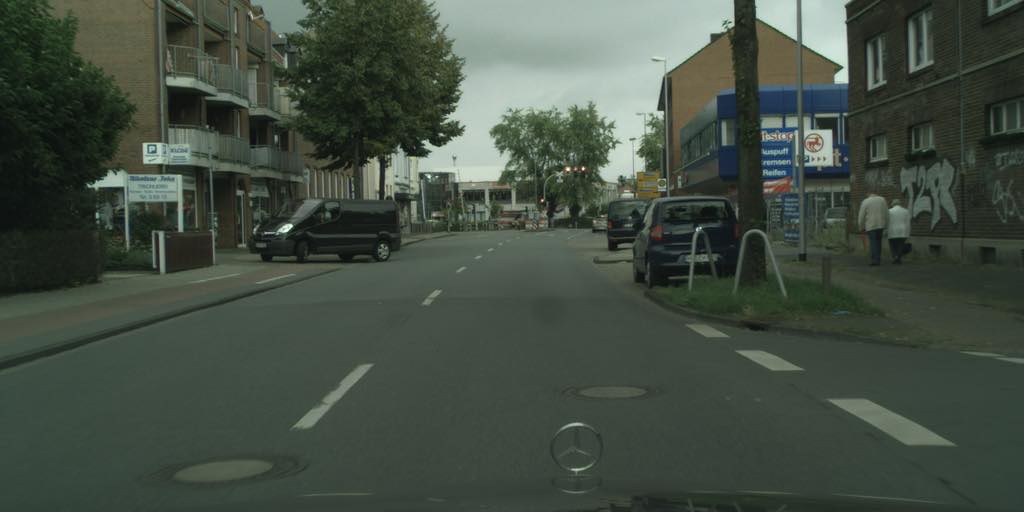} \\
    \rotatebox{90}{~~~~~~~~~~~~~~~Ours} &
    \interpfigcs{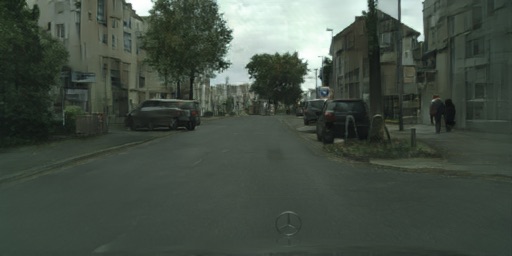} & 
    \interpfigcs{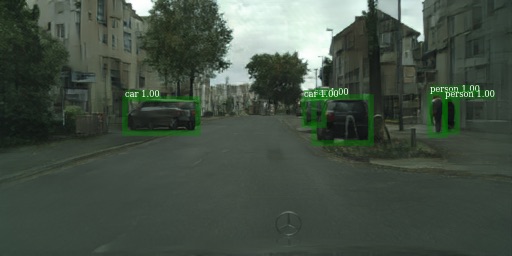} \\
     \rotatebox{90}{~~~~~~~~~~~~~~~SPADE \cite{park2019semantic}} &
    \interpfigcs{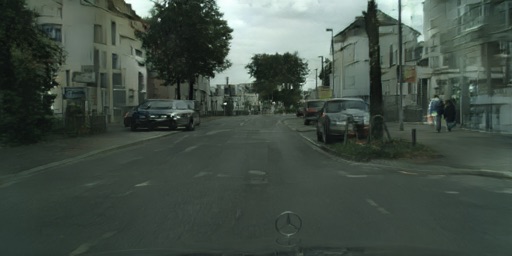}&
     \interpfigcs{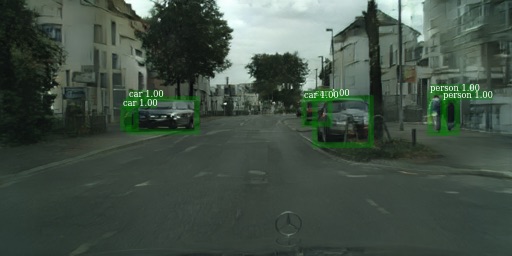}\\
    \rotatebox{90}{~~~~~~~~~~~~~~~CRN \cite{chen2017photographic}}&
    \interpfigcs{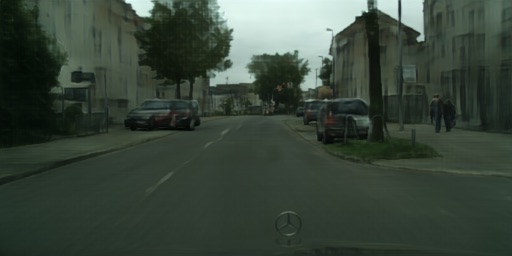}&
    \interpfigcs{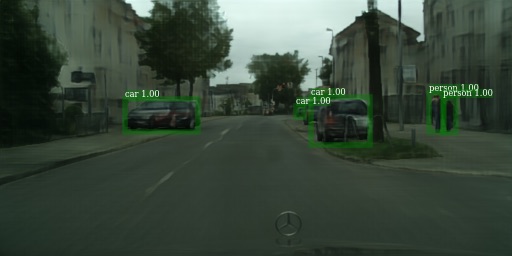}\\
    \rotatebox{90}{~~~~~~~~~~~~~~~SIMS \cite{qi2018semi}}&
    \interpfigcs{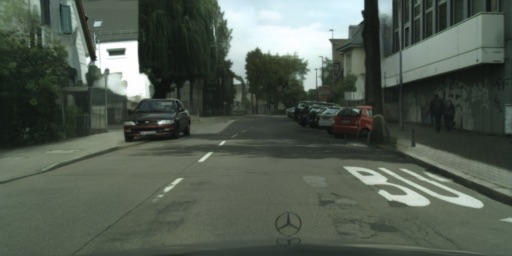}&
     \interpfigcs{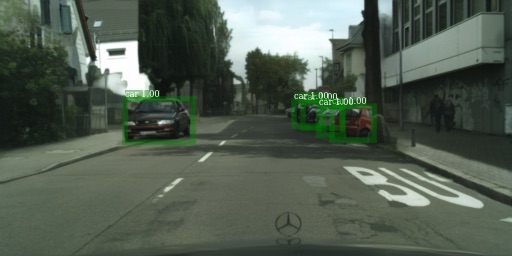}\\
     \end{tabular}
     \label{fig:city_qualitative_5}
\end{figure*}
\end{document}


\title{Supplementary for Panoptic-based Image Synthesis}

\author{Aysegul Dundar \qquad Karan Sapra \qquad Guilin Liu \qquad Andrew Tao \qquad Bryan Catanzaro \\
 NVIDIA Corporation 
}

\maketitle

\newcommand{\interpfigc}[1]{\includegraphics[width=4cm,height=4cm]{#1}}

\begin{figure*}[]
  \centering
   \caption{Additional results on the COCO-Stuff dataset.}
  \setlength\tabcolsep{0.36pt}
  \renewcommand{\arraystretch}{0.15}
  \begin{tabular}{ccccc}
   \rotatebox{90}{~~~~~Panoptic Map \& GT} &
    \interpfigc{Figures/coco/labels/000000507042.png} &
    \interpfigc{Figures/coco/GT/000000507042.jpg} &
    \interpfigc{Figures/coco/labels/000000507975.png} &
    \interpfigc{Figures/coco/GT/000000507975.jpg}  \\
     \rotatebox{90}{~~~~~~~~~~~~~~~Ours} &
    \interpfigc{Figures/coco/ours/000000507042.png} & 
    \interpfigc{Figures/coco/ours_det/000000507042.png}&
    \interpfigc{Figures/coco/ours/000000507975.png} & 
    \interpfigc{Figures/coco/ours_det/000000507975.png} \\
     \rotatebox{90}{~~~~~~~~~~~~~~~SPADE \cite{park2019semantic}} &
    \interpfigc{Figures/coco/spade/000000507042.png}&
     \interpfigc{Figures/coco/spade_det/000000507042.png}&
    \interpfigc{Figures/coco/spade/000000507975.png}&
     \interpfigc{Figures/coco/spade_det/000000507975.png}\\
    \rotatebox{90}{~~~~~~~~~~~~~~~CRN \cite{chen2017photographic}}&
    \interpfigc{Figures/coco/crn/000000507042.png}&
    \interpfigc{Figures/coco/crn_det/000000507042.png} &
    \interpfigc{Figures/coco/crn/000000507975.png}&
    \interpfigc{Figures/coco/crn_det/000000507975.png}\\
     \end{tabular}
     \label{fig:coco_qualitative_1}
\end{figure*}

\begin{figure*}[]
  \centering
  \caption{Additional results on the COCO-Stuff dataset.}
  \setlength\tabcolsep{0.36pt}
  \renewcommand{\arraystretch}{0.15}
  \begin{tabular}{ccccc}
   \rotatebox{90}{~~~~~Panoptic Map \& GT} &
    \interpfigc{Figures/coco/labels/000000516677.png} &
    \interpfigc{Figures/coco/GT/000000516677.jpg} &
    \interpfigc{Figures/coco/labels/000000526706.png} &
    \interpfigc{Figures/coco/GT/000000526706.jpg}  \\
     \rotatebox{90}{~~~~~~~~~~~~~~~Ours} &
    \interpfigc{Figures/coco/ours/000000516677.png} & 
    \interpfigc{Figures/coco/ours_det/000000516677.png}&
    \interpfigc{Figures/coco/ours/000000526706.png} & 
    \interpfigc{Figures/coco/ours_det/000000526706.png} \\
     \rotatebox{90}{~~~~~~~~~~~~~~~SPADE \cite{park2019semantic}} &
    \interpfigc{Figures/coco/spade/000000516677.png}&
     \interpfigc{Figures/coco/spade_det/000000516677.png}&
    \interpfigc{Figures/coco/spade/000000526706.png}&
     \interpfigc{Figures/coco/spade_det/000000526706.png}\\
    \rotatebox{90}{~~~~~~~~~~~~~~~CRN \cite{chen2017photographic}}&
    \interpfigc{Figures/coco/crn/000000516677.png}&
    \interpfigc{Figures/coco/crn_det/000000516677.png} &
    \interpfigc{Figures/coco/crn/000000526706.png}&
    \interpfigc{Figures/coco/crn_det/000000526706.png}\\
     \end{tabular}
     \label{fig:coco_qualitative_2}
\end{figure*}

\begin{figure*}[]
  \centering
\caption{Additional results on the COCO-Stuff dataset showing some more challenging scenario. For example, in figure below detecting two separate instances of elephant is challenging for Faster-RCNN because of the  high overlap of the two instances which leads low fidelity generation of SPADE. However, our proposed changes generate a higher fidelity output, and thus making it to distinct between the two elephant in our synthesized image.}
  \setlength\tabcolsep{0.36pt}
  \renewcommand{\arraystretch}{0.15}
  \begin{tabular}{ccccc}
   \rotatebox{90}{~~~~~Panoptic Map \& GT} &
    \interpfigc{Figures/coco/labels/000000474170.png} &
    \interpfigc{Figures/coco/GT/000000474170.jpg} &
    \interpfigc{Figures/coco/labels/000000314034.png} &
    \interpfigc{Figures/coco/GT/000000314034.jpg}  \\
     \rotatebox{90}{~~~~~~~~~~~~~~~Ours} &
    \interpfigc{Figures/coco/ours/000000474170.png} & 
    \interpfigc{Figures/coco/ours_det/000000474170.png}&
    \interpfigc{Figures/coco/ours/000000314034.png} & 
    \interpfigc{Figures/coco/ours_det/000000314034.png} \\
     \rotatebox{90}{~~~~~~~~~~~~~~~SPADE \cite{park2019semantic}} &
    \interpfigc{Figures/coco/spade/000000474170.png}&
     \interpfigc{Figures/coco/spade_det/000000474170.png}&
    \interpfigc{Figures/coco/spade/000000314034.png}&
     \interpfigc{Figures/coco/spade_det/000000314034.png}\\
    \rotatebox{90}{~~~~~~~~~~~~~~~CRN \cite{chen2017photographic}}&
    \interpfigc{Figures/coco/crn/000000474170.png}&
    \interpfigc{Figures/coco/crn_det/000000474170.png} &
    \interpfigc{Figures/coco/crn/000000314034.png}&
    \interpfigc{Figures/coco/crn_det/000000314034.png}\\
     \end{tabular}
     \label{fig:coco_qualitative_3}
\end{figure*}

\newcommand{\interpfigk}[1]{\includegraphics[width=8cm,height=4cm]{#1}}
\begin{figure*}[]
     \caption{Visual comparison of image synthesis results on Cityscapes dataset. First column shows image synthesis results,  we also provide the bounding box detection predictions from Faster-RCNN in the second column. The cars are partially occluded by poles which create a challenge for the image synthesis methods. CRN generates cars that can be detected by Faster-RCNN but visually look less pleasing. SIMS loosely follows the provided semantic map, and the cars SPADE generates are not distinctive enough to be detected by Faster-RCNN and not of high fidelity.}
  \centering
  \setlength\tabcolsep{0.36pt}
  \renewcommand{\arraystretch}{0.15}
  \begin{tabular}{ccc}
     \rotatebox{90}{~~~~~Panoptic Map \& GT} &
    \interpfigk{Figures/cityscapes/label_ins/frankfurt_000000_011461_leftImg8bit.png} & 
    \interpfigk{Figures/cityscapes/GT/frankfurt_000000_011461_leftImg8bit.png} \\
    \rotatebox{90}{~~~~~~~~~~~~~~~Ours} &
    \interpfigk{Figures/cityscapes/ours/frankfurt_000000_011461_leftImg8bit.png} & 
    \interpfigk{Figures/cityscapes/ours_det/frankfurt_000000_011461_leftImg8bit.png} \\
    \rotatebox{90}{~~~~~~~~~~~~~~~SPADE \cite{park2019semantic}} &
    \interpfigk{Figures/cityscapes/spade/frankfurt_000000_011461_leftImg8bit.png}&
     \interpfigk{Figures/cityscapes/spade_det/frankfurt_000000_011461_leftImg8bit.png}\\
    \rotatebox{90}{~~~~~~~~~~~~~~~CRN \cite{chen2017photographic}}&
    \interpfigk{Figures/cityscapes/crn/frankfurt_000000_011461_leftImg8bit.png}&
    \interpfigk{Figures/cityscapes/crn_det/frankfurt_000000_011461_leftImg8bit.png}\\
    \rotatebox{90}{~~~~~~~~~~~~~~~SIMS \cite{qi2018semi}}&
    \interpfigk{Figures/cityscapes/sims/frankfurt_000000_011461_leftImg8bit.png}&
     \interpfigk{Figures/cityscapes/sims_det/frankfurt_000000_011461_leftImg8bit.png}\\

     \end{tabular}

     \label{fig:city_qualitative_1}
\end{figure*}

\begin{figure*}[]
  \centering
 \caption{Cars generated on the right side of the images present a challenge for the algorithms that only use semantic maps as seen in the images from CRN and SIMS where CRN generates cars with no clear boundaries and SIMS generates four cars while three cars should be present. Thanks to the boundary maps used in SPADE, it can generate the correct number of cars. However, our proposed method along with generating the correct number of car instances, also generates more accurate instances of persons. }
  \setlength\tabcolsep{0.36pt}
  \renewcommand{\arraystretch}{0.15}
  \begin{tabular}{ccc}
     \rotatebox{90}{~~~~~Panoptic Map \& GT} &
    \interpfigk{Figures/cityscapes/label_ins/frankfurt_000001_002646_leftImg8bit.png} &
    \interpfigk{Figures/cityscapes/GT/frankfurt_000001_002646_leftImg8bit.png} \\
    \rotatebox{90}{~~~~~~~~~~~~~~~Ours} &
    \interpfigk{Figures/cityscapes/ours/frankfurt_000001_002646_leftImg8bit.png} & 
    \interpfigk{Figures/cityscapes/ours_det/frankfurt_000001_002646_leftImg8bit.png} \\
     \rotatebox{90}{~~~~~~~~~~~~~~~SPADE \cite{park2019semantic}} &
    \interpfigk{Figures/cityscapes/spade/frankfurt_000001_002646_leftImg8bit.png}&
     \interpfigk{Figures/cityscapes/spade_det/frankfurt_000001_002646_leftImg8bit.png}\\
    \rotatebox{90}{~~~~~~~~~~~~~~~CRN \cite{chen2017photographic}}&
    \interpfigk{Figures/cityscapes/crn/frankfurt_000001_002646_leftImg8bit.png}&
    \interpfigk{Figures/cityscapes/crn_det/frankfurt_000001_002646_leftImg8bit.png}\\
    \rotatebox{90}{~~~~~~~~~~~~~~~SIMS \cite{qi2018semi}}&
    \interpfigk{Figures/cityscapes/sims/frankfurt_000001_002646_leftImg8bit.png}&
     \interpfigk{Figures/cityscapes/sims_det/frankfurt_000001_002646_leftImg8bit.png}\\
     \end{tabular}

     \label{fig:city_qualitative_2}
\end{figure*}

\begin{figure*}[]
  \centering
  \caption{Additional results on the Cityscapes dataset.}
  \setlength\tabcolsep{0.36pt}
  \renewcommand{\arraystretch}{0.15}
  \begin{tabular}{ccc}
     \rotatebox{90}{~~~~~Panoptic Map \& GT} &
    \interpfigk{Figures/cityscapes/label_ins/munster_000122_000019_leftImg8bit.png} &
    \interpfigk{Figures/cityscapes/GT/munster_000122_000019_leftImg8bit.png} \\
    \rotatebox{90}{~~~~~~~~~~~~~~~Ours} &
    \interpfigk{Figures/cityscapes/ours/munster_000122_000019_leftImg8bit.png} & 
    \interpfigk{Figures/cityscapes/ours_det/munster_000122_000019_leftImg8bit.png} \\
     \rotatebox{90}{~~~~~~~~~~~~~~~SPADE \cite{park2019semantic}} &
    \interpfigk{Figures/cityscapes/spade/munster_000122_000019_leftImg8bit.png}&
     \interpfigk{Figures/cityscapes/spade_det/munster_000122_000019_leftImg8bit.png}\\
    \rotatebox{90}{~~~~~~~~~~~~~~~CRN \cite{chen2017photographic}}&
    \interpfigk{Figures/cityscapes/crn/munster_000122_000019_leftImg8bit.png}&
    \interpfigk{Figures/cityscapes/crn_det/munster_000122_000019_leftImg8bit.png}\\
    \rotatebox{90}{~~~~~~~~~~~~~~~SIMS \cite{qi2018semi}}&
    \interpfigk{Figures/cityscapes/sims/munster_000122_000019_leftImg8bit.png}&
     \interpfigk{Figures/cityscapes/sims_det/munster_000122_000019_leftImg8bit.png}\\
     \end{tabular}
     \label{fig:city_qualitative_3}
\end{figure*}

\begin{figure*}[]
  \centering
     \caption{Additional results on the Cityscapes dataset.}
  \setlength\tabcolsep{0.36pt}
  \renewcommand{\arraystretch}{0.15}
  \begin{tabular}{ccc}
     \rotatebox{90}{~~~~~Panoptic Map \& GT} &
    \interpfigk{Figures/cityscapes/label_ins/munster_000118_000019_leftImg8bit.png} &
    \interpfigk{Figures/cityscapes/GT/munster_000118_000019_leftImg8bit.png} \\
    \rotatebox{90}{~~~~~~~~~~~~~~~Ours} &
    \interpfigk{Figures/cityscapes/ours/munster_000118_000019_leftImg8bit.png} & 
    \interpfigk{Figures/cityscapes/ours_det/munster_000118_000019_leftImg8bit.png} \\
     \rotatebox{90}{~~~~~~~~~~~~~~~SPADE \cite{park2019semantic}} &
    \interpfigk{Figures/cityscapes/spade/munster_000118_000019_leftImg8bit.png}&
     \interpfigk{Figures/cityscapes/spade_det/munster_000118_000019_leftImg8bit.png}\\
    \rotatebox{90}{~~~~~~~~~~~~~~~CRN \cite{chen2017photographic}}&
    \interpfigk{Figures/cityscapes/crn/munster_000118_000019_leftImg8bit.png}&
    \interpfigk{Figures/cityscapes/crn_det/munster_000118_000019_leftImg8bit.png}\\
    \rotatebox{90}{~~~~~~~~~~~~~~~SIMS \cite{qi2018semi}}&
    \interpfigk{Figures/cityscapes/sims/munster_000118_000019_leftImg8bit.png}&
     \interpfigk{Figures/cityscapes/sims_det/munster_000118_000019_leftImg8bit.png}\\
     \end{tabular}
     \label{fig:city_qualitative_4}
\end{figure*}

\begin{figure*}[]
  \centering
  \caption{Additional results on the Cityscapes dataset.}
  \setlength\tabcolsep{0.36pt}
  \renewcommand{\arraystretch}{0.15}
  \begin{tabular}{ccc}
     \rotatebox{90}{~~~~~Panoptic Map \& GT} &
    \interpfigk{Figures/cityscapes/label_ins/munster_000113_000019_leftImg8bit.png} &
    \interpfigk{Figures/cityscapes/GT/munster_000113_000019_leftImg8bit.png} \\
    \rotatebox{90}{~~~~~~~~~~~~~~~Ours} &
    \interpfigk{Figures/cityscapes/ours/munster_000113_000019_leftImg8bit.png} & 
    \interpfigk{Figures/cityscapes/ours_det/munster_000113_000019_leftImg8bit.png} \\
     \rotatebox{90}{~~~~~~~~~~~~~~~SPADE \cite{park2019semantic}} &
    \interpfigk{Figures/cityscapes/spade/munster_000113_000019_leftImg8bit.png}&
     \interpfigk{Figures/cityscapes/spade_det/munster_000113_000019_leftImg8bit.png}\\
    \rotatebox{90}{~~~~~~~~~~~~~~~CRN \cite{chen2017photographic}}&
    \interpfigk{Figures/cityscapes/crn/munster_000113_000019_leftImg8bit.png}&
    \interpfigk{Figures/cityscapes/crn_det/munster_000113_000019_leftImg8bit.png}\\
    \rotatebox{90}{~~~~~~~~~~~~~~~SIMS \cite{qi2018semi}}&
    \interpfigk{Figures/cityscapes/sims/munster_000113_000019_leftImg8bit.png}&
     \interpfigk{Figures/cityscapes/sims_det/munster_000113_000019_leftImg8bit.png}\\
     \end{tabular}
     \label{fig:city_qualitative_5}
\end{figure*}

\newpage

{\small
\bibliographystyle{ieee_fullname}
\bibliography{egbib}
}